\documentclass[10pt,twocolumn,letterpaper]{article}

\usepackage[pagenumbers]{iccv} %
\usepackage[accsupp]{axessibility}

\usepackage{multirow} 
\PassOptionsToPackage{HTML}{xcolor}
\usepackage{wasysym}
\usepackage{bm}

\usepackage{mydefs}

\definecolor{iccvblue}{rgb}{0.21,0.49,0.74}
\usepackage[pagebackref,breaklinks,colorlinks,allcolors=iccvblue]{hyperref}

\def\paperID{501} %
\def\confName{ICCV}
\def\confYear{2025}

\title{Is Tracking really more challenging in First Person Egocentric Vision?}

\author{ 
Matteo Dunnhofer$^{1,2}$
\quad\ Zaira Manigrasso$^{1}$
\quad\ Christian Micheloni$^{1}$ 
\\
\\
$^{1}$University of Udine, Italy \quad\
$^{2}$York University, Canada 
}

\begin{document}

\twocolumn[{%
\renewcommand\twocolumn[1][]{#1}%
\maketitle
\vspace{-0.9cm}
\begin{center}
    \centering
    \captionsetup{type=figure}
    \includegraphics[width=\linewidth]{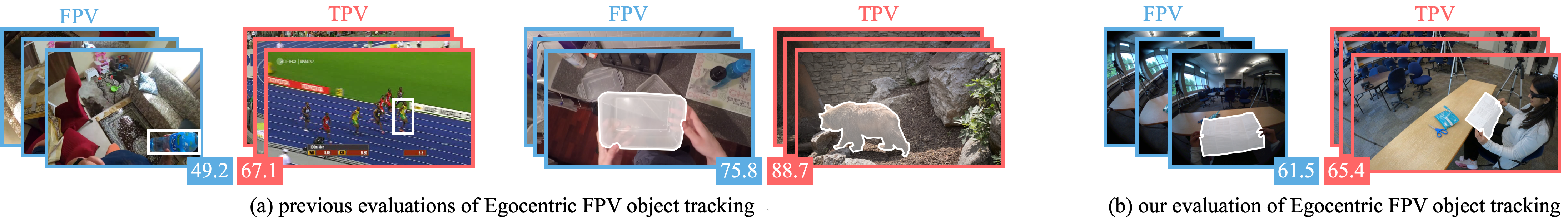}
    \captionof{figure}{\textbf{We isolate the real impact of first person vision (FPV) in  object tracking evaluation.} (a) Previous benchmarks claimed the challenges of tracking in FPV by comparing algorithm performance on third person vision (TPV) datasets having mismatched object categories and contexts. This inconsistency makes it difficult to isolate the true impact of the FPV viewpoint.
    (b) This paper addresses this problem by directly evaluating object tracking in synchronized FPV and TPV videos of the same human-object activity. 
    }
    \label{fig:idea}
\end{center}%
}]

\maketitle

\begin{abstract}
Visual object tracking and segmentation are becoming fundamental tasks for understanding human activities in egocentric vision. Recent research has benchmarked state-of-the-art methods and concluded that first person egocentric vision presents  challenges compared to  previously studied domains. However, these claims are based on evaluations conducted across significantly different scenarios. Many of the challenging characteristics attributed to egocentric vision are also present in third person videos of human-object activities.
This raises a critical question: how much of the observed performance drop stems from the unique first person viewpoint inherent to egocentric vision versus the domain of human-object activities? To address this question, we introduce a new benchmark study designed to disentangle such factors. Our evaluation strategy enables a more precise separation of challenges related to the first person perspective from those linked to the broader domain of human-object activity understanding. By doing so, we provide deeper insights into the true sources of difficulty in egocentric tracking and segmentation, facilitating more targeted advancements on this task.
\end{abstract}

\section{Introduction}
\label{sec:intro}

Recent studies \cite{dunnhofer2023visual, tang2024egotracks,darkhalil2022epic} have highlighted the importance of visual object tracking and segmentation \cite{wu2013online,perazzi2016benchmark, kristan2023first} for understanding activities in egocentric first person vision (FPV). The instance-based localization of objects within 2D egocentric video frames is an intermediate task for high-level vision-based assistive systems, including 3D object localization \cite{zhao2024instance,hao2024ego3dt,plizzari2024spatial}, episodic memory  \cite{grauman2022ego4d,manigrasso2024online}, and hand-object interaction understanding \cite{dunnhofer2023visual,darkhalil2022epic,goletto2024amego}.

Despite its promising applications, 2D visual object tracking remains an open problem in egocentric vision. This limitation stems  from the early stage of research in this domain, where the unique challenges of this problem are not yet fully understood.
Indeed, recent benchmarking studies \cite{dunnhofer2023visual,tang2024egotracks,darkhalil2022epic} scratched the surface of the problem by evaluating generalistic state-of-the-art methodologies on a set of egocentric videos to assess their effectiveness. These studies concluded that object tracking in FPV poses greater challenges than the more general settings found in established benchmarks \cite{LaSOT,GOT10k,kristan2023first,perazzi2016benchmark,xu2018youtube}, and that current tracking models lack the necessary mechanisms to effectively handle these difficulties.

A significant issue with these claims lies in the fact that the supporting comparisons have been conducted in vastly different domains, with limited alignment in terms of object categories, motion patterns, interactions, and contextual characteristics. For example, the TREK-150 \cite{dunnhofer2023visual} and VISOR \cite{darkhalil2022epic} benchmarks were introduced to evaluate visual object tracking (VOT) and video object segmentation (VOS) algorithms exclusively within kitchen environments, whereas the EgoTracks \cite{tang2024egotracks} benchmark, although covering a broader range of object categories and scenarios, still lacks alignment with the objects  and contexts typically featured in widely used benchmarks \cite{wu2013online,perazzi2016benchmark,kristan2023first,LaSOT,GOT10k,TrackingNet}. Many works in egocentric vision \cite{dunnhofer2023visual,tang2024egotracks, grauman2022ego4d,EgoExo4D} refer to these datasets  as representations of third person vision (TPV) scenarios, where target objects are passively observed from an external perspective that lacks embodiment of the camera wearer. Consequently, previous benchmarks of egocentric object tracking have concluded that FPV is more challenging compared to TPV.

However, many characteristics often attributed to FPV are also present in videos of the broader domain of human activity understanding \cite{EgoExo4D}. Indeed, videos of humans performing activities involving specific types of objects --  moving, hiding, or transforming them -- are not exclusively captured from an FPV perspective, but can also be recorded by stationary cameras in the environment that passively observe the scene from a TPV point of view.
Given these considerations, one could argue that the decline in performance of object tracking models is not inherently due to the FPV viewpoint itself but rather to the domain of human activity understanding, which is fundamentally different from the domains on which these models are typically trained.
This misalignment raises the question reflected in the title of this paper: is the challenge of FPV object tracking primarily due to unique viewpoint characteristics, or is it because it represents a specific example of human-object acitivity understanding that is not adequately captured in existing object tracking datasets?

We present a novel study to give  answers to such a question, and we test the claim wether FPV is really different and more challenging for object tracking.
As depicted in Fig. \ref{fig:idea}, we do this by contrasting FPV tracking performance against TPV tracking performance, in synchronized videos capturing the same human-object activity from the two perspectives. Our evaluation protocol ensures frame-level synchronization between the two perspectives and aligned object annotations across both views. This setup enables a consistent comparison of  tracking performance that isolates the viewpoint bias within the domain of human activity understanding videos.
We implement this experimental scheme over a recent dataset \cite{EgoExo4D} and compile a new set of data. Overall, these contributions give birth to a new tracking benchmark that we name \benchmarkname, because it provides multiple \emph{vistas} of the same scene.

We use \benchmarkname\ to evaluate state-of-the-art VOT and VOS methods, as recently merged in the visual object tracking segmentation (VOTS) task \cite{kristan2020eighth,kristan2023first, kristan2024second}. Despite differences in target state, these methods share the core goal of continuous object instance localization \cite{kristan2023first, dunnhofer2023visual,darkhalil2022epic}.
The unified evaluation framework highlights the unique challenges FPV poses to current 2D object tracking methods, revealing their limitations in egocentric vision and informing directions for future improvement. We believe our in-depth findings will enhance the understanding and development of VOTS in both FPV and TPV, and contribute to the field of computer vision for human activity understanding more in general.
Code, data, and results, are available at \datasetlink.

\section{Related Work}
\label{sec:rw}

\paragraph{Benchmarking tracking in egocentric FPV.}

TREK-150  \cite{dunnhofer2021first,dunnhofer2023visual} was the first systematic effort to evaluate state-of-the-art bounding-box trackers in egocentric FPV. It introduced a dataset along with evaluation protocols and metrics relevant for egocentric downstream tasks.
Building on this, EgoTracks \cite{tang2024egotracks} expanded the analysis to a larger dataset with longer video durations, providing a benchmark for egocentric long-term VOT.
Parallely, the VISOR dataset \cite{darkhalil2022epic} was introduced to evaluate semi-supervised VOS for segmenting hands and active objects in hand-object interaction understanding. The VOST benchmark \cite{tokmakov2023breaking} evaluated VOS in egocentric videos where objects undergo structural transformations over time.
More recently, the growing development in point tracking \cite{harley2022particle,zheng2023pointodyssey,karaev2024cotracker} was studied in egovision with the release of the EgoPoints benchmark \cite{darkhalil2024egopoints}.

The aforementioned studies claimed FPV more challenging than TPV and other domains. However, these conclusions were drawn from comparisons across different contexts, dataset sizes, and training conditions, making it difficult to isolate FPV-specific challenges.
This paper addresses these limitations by separating FPV challenges from those of human activity videos. We present a novel benchmark study that uses synchronized FPV and TPV recordings of human activities for a direct viewpoint comparison.

\paragraph{Benchmarking generic VOT, VOS, and VOTS.}

Evaluating generic-object bounding-box-based tracking algorithms has been a cornerstone of computer vision research over the past decade.
The OTB benchmark \cite{wu2013online,wu2015} was among the first systematic evaluations, featuring 100 annotated videos with diverse targets and variability factors.
Meanwhile, the VOT Challenge series \cite{kristan2015visual,kristan2018sixth,kristan2019seventh} became a key benchmark for causal VOT algorithms, providing standardized protocols and metrics for assessing trackers in short-term, long-term, and real-time conditions.
Other  benchmarks assessed VOT based on specific video characteristics or modalities: TC-128 \cite{liang2015encoding} examined color information; NfS \cite{kiani2017need} analyzed high frame rates; and CDTB \cite{lukezic2019cdtb} evaluated the integration of RGB data with depth. Other specialized benchmarks focused on extreme video conditions caused by  severe weather \cite{noman2022avist}, night-time \cite{liu2024nt}, or camera shot-cuts \cite{dunnhofer2024tracking}.
The advent of deep learning necessitated larger and more diverse datasets to train and evaluate VOT models effectively. Benchmarks such as TrackingNet \cite{TrackingNet}, LaSOT \cite{LaSOT}, and GOT-10k \cite{GOT10k} addressed this need by providing large-scale datasets comprising thousands of annotated videos divided in training and test sets.

Alongside box-based VOT, the computer vision community has focused on VOS, where target states are represented by segmentation masks.
The DAVIS benchmark \cite{perazzi2016benchmark} standardized VOS evaluation with mask annotations and metrics for single-object tracking in short, high-quality videos. DAVIS 2017 \cite{pont2017} extended this to multi-object segmentation, requiring algorithms to track multiple instances.
Large-scale benchmarks have further advanced VOS. YouTube-VOS \cite{xu2018youtube} introduced over 4,000 videos across over 100 categories, including unseen objects to test generalization. MOSE \cite{ding2023mose} focused on VOS in dynamic real-world settings, emphasizing frame-to-frame consistency. LVOS \cite{hong2023lvos} studied long-term VOS, providing extended sequences to evaluate temporal consistency and robustness.

More recent studies \cite{kristan2020eighth,kristan2023first,kristan2024second} introduced the VOTS benchmark, unifying the tasks of VOT and VOS by framing them as variations of the same core problem: tracking a generic object specified in the first frame of a video. 
These efforts underscored the importance of developing unified methodologies to address pixel-precise object tracking.
Building on this foundation, our work offers a comprehensive evaluation of both VOT and VOS approaches within a unified benchmark framework.

Unlike all these studies using single videos, this paper introduces a novel  study that employs an evaluation protocol applied to synchronized videos capturing the same scene from different perspectives. This approach aims to isolate the impact of the video viewpoint on tracking. To our knowledge, this is the first study to investigate potential biases in VOTS methods based on the camera’s perspective.

\paragraph{Benchmarking egocentric FPV.}
Benchmarking egocentric vision tasks has evolved significantly in the past years, with datasets progressively capturing more complex problems and scenarios. Early benchmarks like ADL \cite{pirsiavash2012detecting} and GTEA \cite{li2015delving} focused on evaluating algorithms in short video sequences of daily activities, offering annotations for action recognition and object manipulation. 
EPIC-Kitchens \cite{damen2018scaling} advanced the field with a large-scale dataset of kitchen activities, annotated for action recognition, anticipation and retrieval, and object detection. 
Ego4D \cite{grauman2022ego4d} expanded the scope and scale, providing 3,670 hours of egocentric videos in diverse world-wide scenarios. It defined five core tasks: episodic memory, hand-object interaction understanding, social interaction understanding, and activity forecasting. 

The more recent EgoExo4D dataset \cite{EgoExo4D} uniquely combines synchronized egocentric, i.e. FPV, and exocentric, i.e. TPV, video streams to benchmark cross-view video understanding. Beyond popular egocentric vision tasks \cite{grauman2022ego4d}, it introduces the Correspondence task, which requires locating objects simultaneously in FPV and TPV videos after initialization in one of the views.
While related, our research objective differs. We use temporal alignment between FPV and TPV to examine how viewpoint differences in object and environment appearance affect tracking. This allows us to test the claim that FPV is inherently more challenging than TPV for the performance of VOTS algorithms.

\section{The \benchmarkname\ Benchmark}
This paper aims to understand if FPV is truly a distinct and challenging domain for VOTS. Specifically, we investigate  whether differences in the behavior of algorithms arise due to the FPV viewpoint by comparing against synchronized TPV videos capturing the same scene. To this end, we introduce \benchmarkname, a comprehensive benchmark with tailored evaluation protocols, metrics, and annotated videos.\footnote{Refer to \ref{appendix:benchmark} of the Supp. Mat. for further details.}

\subsection{Evaluation}
\label{sec:prot}

\paragraph{Synchronized One-Pass Evaluation Protocol.}
To determine the impact of FPV on VOTS, we compare FPV tracking performance against TPV tracking performance. To this end, we design an evaluation protocol that uses two synchronized annotated videos of equal length, capturing the same scene from FPV and TPV. We extend the standard one-pass evaluation (OPE) \cite{wu2013online,wu2015,perazzi2016benchmark}, also referred to semi-supervised protocol \cite{perazzi2016benchmark}, to work with synchronized videos, introducing a new protocol called SOPE — synchronized one-pass evaluation.
SOPE assumes the availability of two annotated videos as pairs 
$\videofpv = (\framesfpv, \annos^{\fpv})     \text{ and } \videotpv = (\framestpv, \annos^{\tpv})$,
where
$\framesfpv = \{\frame_t^{\fpv}\}_{t=0}^{T-1}, \: \framestpv = \{\frame_t^{\tpv}\}_{t=0}^{T-1}$
are equal-length sequences of $T$ RGB frames that have paired annotations 
$\annos^{\fpv} = \{\anno_t^{\fpv}\}_{t=0}^{T-1}, \: \annos^{\tpv} = \{\anno_t^{\tpv}\}_{t=0}^{T-1}$.
Annotations can be bounding-boxes
$\bbox_t^{\fpv},\bbox_t^{\tpv}$ or segmentation masks $\mask_t^{\fpv},\mask_t^{\tpv}$ representing the target object state in the two videos.
$\anno_t^{\fpv}$ and $\anno_t^{\tpv}$ may be empty for certain time steps $t$, but we put the constraint that they must be non-empty for the same subset of time steps.
With this data, SOPE runs a tracker separately on each annotated video $\videopov = (\framespov, \annos^{\pov}), \pov \in \{\fpv, \tpv\}$. The algorithm is first initialized with $\frame_0^{\pov}$ and either $\bbox_0^{\pov}$ or $\mask_0^{\pov}$ depending on the target representation accepted by the tracker. The algorithm then has to track the object in subsequent frames $\frame_t^{\pov}, 0 < t \leq T-1$, in an online manner, outputting predictions $\pred_t^{\pov}$ as either boxes $\widehat{\bbox}_t^{\pov}$ or masks $\widehat{\mask}_t^{\pov}$ based on its output target representation. These outputs are recorded for later performance scoring.
We iterate the described two steps for all the sequences present in an evaluation dataset $\testset = \big\{(\video_i^{\fpv}, \video_i^{\tpv})\big\}_{i=0}^{N-1}$  of $N$ synchronized videos, in order to determine the average tracking performance in FPV and TPV, and their difference.

\begin{table*}
\caption{\textbf{Comparison of the \benchmarkname benchmark's dataset with other egocentric tracking-related datasets.} The dataset used in our study is comparable in scale to previous real-world datasets designed for evaluating egocentric object tracking. However, unlike these, we include synchronized TPV videos to isolate the influence of the visual characteristics unique to FPV videos.}
    \label{tab:stats}
    \centering

    \tblalternaterowcolors
    
    \fontsize{9}{8}\selectfont

    \setlength\tabcolsep{.17cm}
    
    \begin{tabular}{l  l | c c c c c c c c }
    \toprule

    Benchmark & Tracking focus & Sequences & Frames & Duration (m) & Avg.  Duration (m) & Train-Test & Viewpoint  \\

    \midrule

        TREK-150 \cite{dunnhofer2023visual} & hand-object interaction & 
         150 & 97,000 & 27 & 0.2 & \xmark & FPV  \\
        
        EgoTracks  \cite{tang2024egotracks} & long-term object tracking & 22028 & 10,274,400 & 36,174 & 6.1 & \cmark & FPV  \\

        IT3DEgo \cite{zhao2024instance} & 3D object tracking  & 240 & 500,000 & 250 & 5 & \xmark & FPV  \\

        VISOR \cite{darkhalil2022epic} & object manipulation & 7836 & 50,700 & 2160 & 0.1 & \cmark & FPV  \\

        VOST \cite{tokmakov2023breaking} & object transformation & 713 & 75,547 & 143 & 0.2 & \cmark & FPV  \\

        EgoPoints \cite{darkhalil2024egopoints} & point tracking & 517 & 264,187 & 73 & 0.8 & \xmark & FPV  \\        
        
        \midrule
        
        \rowcolor{white} & & 2375 & 2,190,480 & 7302 & 3.1 &  \cmark & FPV  \\
        \multirow{-2}{*}{\benchmarkname} & \multirow{-2}{*}{viewpoint impact} & 2375 & 2,190,480  & 7302 & 3.1 & \cmark & TPV  \\

        \bottomrule
    \end{tabular}
    
\end{table*}

\paragraph{Measures.}
To quantify the effect of FPV on a tracker, we measure the difference between its FPV and TPV behavior. We use standard tracking performance metrics along with methods to measure their differences.
Specifically, for each pair $i$ of annotated videos in $\testset$, we define  
    $s_{\sigma,i}^{\pov} = \sigma\big(\{\pred_t^{\pov} \}, \{ \anno_t^{\pov} \}\big)$
the score given by a function $\sigma(\cdot, \cdot)$ that computes the discrepancy between tracker predictions and ground-truth annotations.  
To implement $\sigma(\cdot, \cdot)$, we use the IoU-based Area Under the Curve (AUC) \cite{wu2013online}, the Normalized Precision Score (NPS) \cite{LaSOT}, and the Generalized Success Robustness (GSR) \cite{dunnhofer2023visual}, all ranging in $[0, 100]$. These metrics are chosen for their prior use in FPV benchmarking \cite{dunnhofer2023visual} and their status as standard metrics in generic VOTS evaluation \cite{wu2013online,kristan2020eighth,LaSOT,kristan2023first}.   
The performance difference is quantified as the mean weighted signed difference
\small
\begin{align}
\label{eq:diff}
\Delta_{\sigma} = \frac{1} {\omega_0 + \cdots + \omega_{N-1}}\sum_{i=0}^{N-1} (s_{\sigma,i}^{\fpv} - s_{\sigma,i}^{\tpv}) \cdot \omega_i, \omega_i = |\annos^{pov}_i|.
\end{align}  
\normalsize
This is a simple, interpretable, and widely used method for quantifying differences, offering a direct measure of improvement or degradation. Its linearity and sensitivity to small changes make it effective for model comparisons \cite{willmott1982some}.
We apply this approach to AUC, NPS, and GSR, giving $\Delta_{\textbf{AUC}}, \Delta_{\textbf{NPS}}$, $\Delta_{\textbf{GSR}}$. 
When reported as test-set-averaged AUC-$\pov$, NPS-$\pov$, GSR-$\pov$, the $s_{\sigma,i}^{\pov}$ are also weighted by the annotation length $\omega_i$, following \cite{kristan2016novel}. As detailed in \ref{appendix:metrics} of the Supp. Mat., the weighting ensures consistent evaluation when tracking behavior varies in the two views.

To ensure $\Delta_{\sigma}$ reflects only the viewpoint difference, the following additional constraint must be enforced: $\frame_t^{\fpv}$ and $\frame_t^{\tpv}$ must capture the same scene, with $\frame_t^{\fpv}$ observed from an FPV viewpoint \cite{grauman2022ego4d} and $\frame_t^{\tpv}$ from a TPV viewpoint \cite{EgoExo4D}. By maintaining shared scene observation, synchronized videos, equal video lengths, and the same number of annotations, we ensure that $\Delta_{\sigma}$ captures only the differences in tracker performance due to the viewpoint change.

\begin{figure*}[t]
    \centering
    \includegraphics[width=1\linewidth]{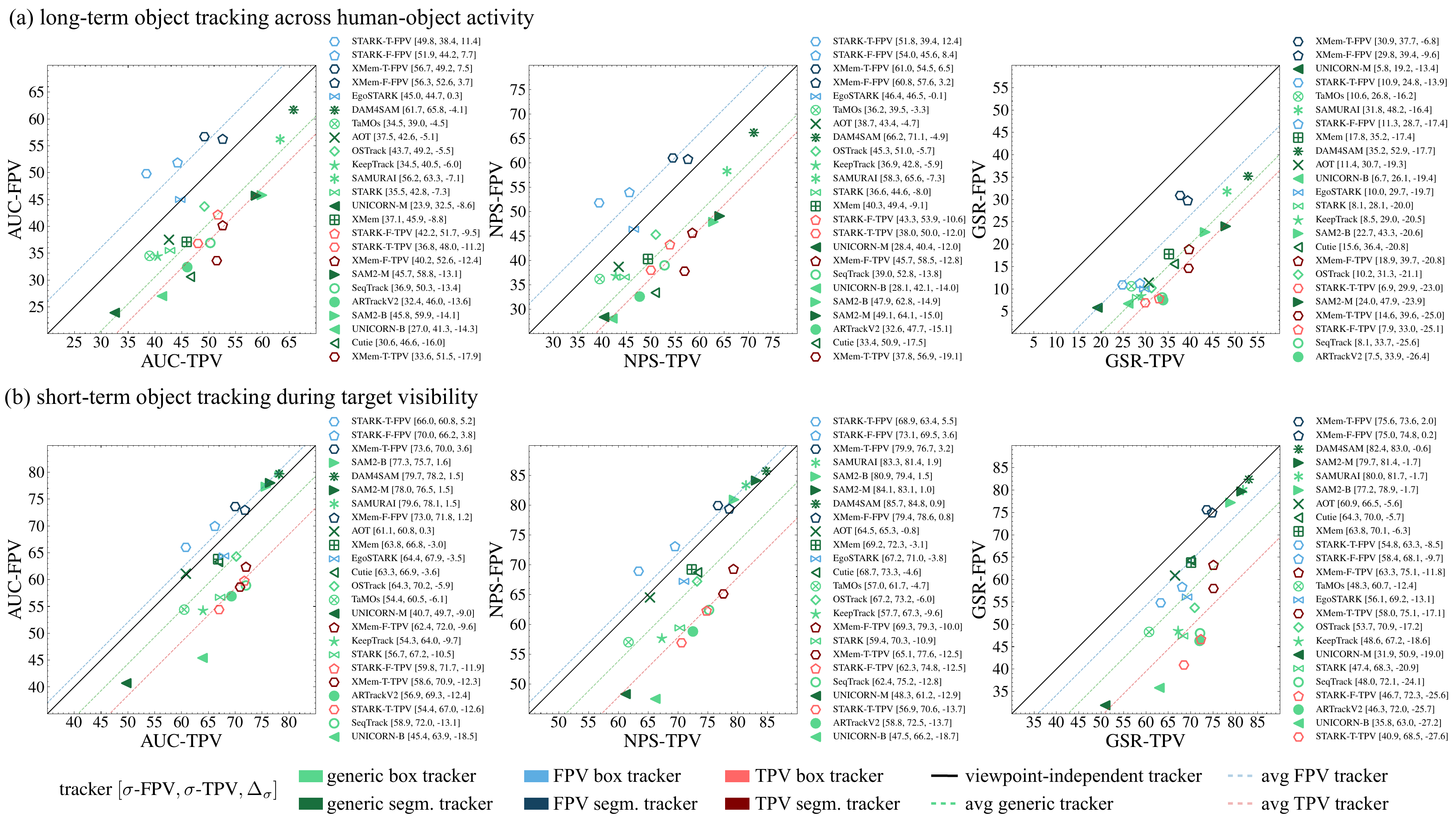}
    \caption{\textbf{\plotname Plots for FPV and TPV.} (a) When tasked with tracking a target object throughout an entire human-object activity, generic object trackers perform better in TPV, confirming the challenges posed by FPV. FPV-specific trackers improve FPV performance and lose accuracy in TPV, while TPV-trained trackers perform better in TPV and sacrifice FPV accuracy. (b) When tracking a target object for a short period of visibility, viewpoint bias is reduced for generic and FPV object trackers. However, TPV-trained trackers show a clear bias towards TPV. Scores in brackets are ordered in descending order by $\Delta_{\sigma}$. See Tab. \ref{tab:generictrackerlt} of the Supp. Mat. for a better readability of the scores.}
    \label{fig:scatterplotlt}
\end{figure*}

\paragraph{\plotname Plot.} 
We introduce the \plotname Plot, a novel method to visualize performance differences between FPV and TPV. This scatter plot displays the average %
$\sigma$-FPV score 
on the vertical axis and the %
$\sigma$-TPV score 
on the horizontal axis, with each tracker represented by a point at coordinates 
($\sigma$-TPV, $\sigma$-FPV). 
A solid diagonal line represents view-independent performance. Trackers on the line exhibit no bias, and those off the diagonal show bias proportional to their mean weighted signed difference. 
A robust VOTS method will appear in the top-right section, indicating high, view-independent behavior. This visualization can be applied to various metrics, as shown for AUC, NPS, and GSR in Fig. \ref{fig:scatterplotlt}. %

\begin{figure*}[th]
  \centering
  \includegraphics[width=\linewidth]{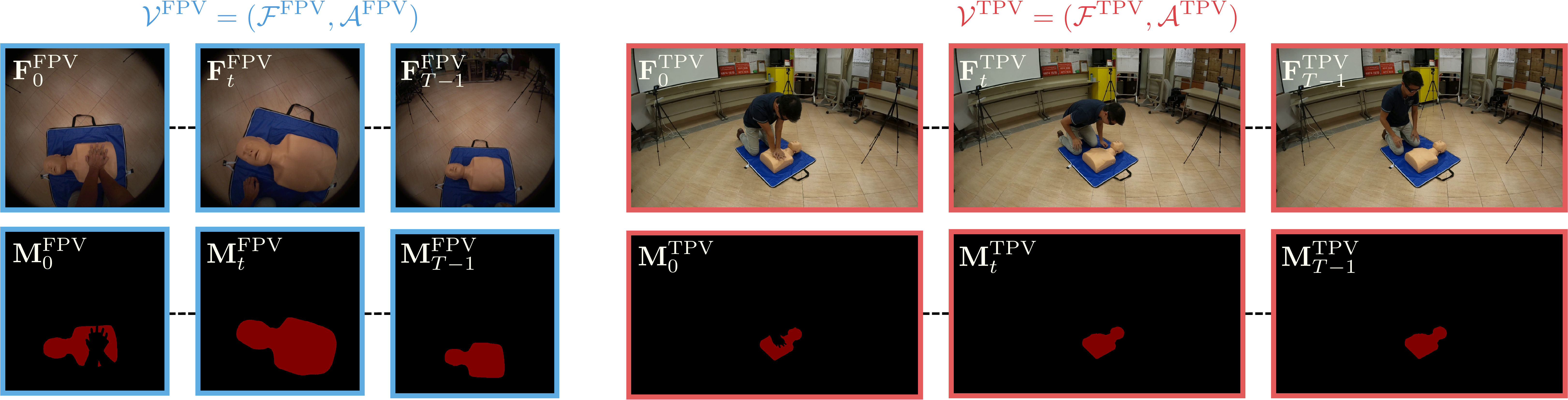}
  \caption{\textbf{Example of a synchronized FPV and TPV video.}
  The \benchmarkname\ benchmark dataset is composed of pairs of annotated videos $\videofpv, \videotpv$ where FPV frames $\frames^{\fpv} = \{ \frame_t^{\fpv} \}_{t=0}^{T-1}$ are synchronized to TPV frames $\frames^{\tpv} = \{ \frame_t^{\tpv} \}_{t=0}^{T-1}$ and both are annotated with the ground-truth state of a target object. In this example, annotations are segmentation masks $\annos^{\fpv} = \{ \mask_t^{\fpv} \}_{t=0}^{T-1}, \annos^{\tpv} = \{ \mask_t^{\tpv} \}_{t=0}^{T-1}$.}
  \label{fig:example}
\end{figure*}

\subsection{Data}
\label{sec:data}
The \benchmarkname\ benchmark includes data to implement the SOPE protocol, ensuring compliance with its constraints for viewpoint impact isolation. It features a test set $\testset$ of $N =$ 544 annotated FPV-TPV video pairs and a training set $\trainset = {(\videofpv_j, \videotpv_j)}_{j=0}^{M-1}$ of $M =$ 1831 annotated FPV-TPV video pairs, all adhering to the contstraints outlined before. The test set is used for tracker evaluation, while the training set is employed to train deep learning-based models and assess the impact of viewpoint during learning. Dataset characteristics (TRAIN+TEST) are summarized in Tab. \ref{tab:stats}, highlighting that it is the only dataset offering synchronized FPV-TPV object tracks with a scale and temporal length that balances variability and evaluation time.

\paragraph{Video Collection and Annotation.}
We constructed the dataset using videos and annotations from the EgoExo4D dataset \cite{EgoExo4D}, the largest and most diverse resource offering synchronized FPV and TPV captures of the same human activity (e.g. cooking, sports, instrument playing, item repair, etc.). Starting with the original annotations for the Relations benchmark \cite{EgoExo4D}, we filtered segmentation masks based on the previuosly defined constraints. We selected instance-level sequences with an annotation rate of 1 frame per second, as recommended for reliable tracking evaluation \cite{lukezivc2020performance}, and retained only FPV and TPV sequences with aligned mask annotations. To minimize storage without compromising quality, all frames and masks were resized to a minimum dimension of 720 pixels, and videos were saved at 5 FPS \cite{tang2024egotracks}.
The sequences in $\testset$ were derived from EgoExo4D's validation set, resulting in 544 single-object tracks per viewpoint across 140 video clips and 140 object categories. The same procedure was applied to generate $\trainset$. $\testset$ contains 550,794 frames with 60,986 mask annotations per viewpoint, while $\trainset$ includes 1,639,686 frames with 190,571 mask annotations per viewpoint. The average sequence length is 922 frames (approximately 3.1 minutes), representing a long-term tracking scenario \cite{tang2024egotracks}, where the object is tracked continuously throughout the activity.
An example of sequence is shown in Fig. \ref{fig:example}.
In addition, we included frame-level attributes to capture the visual variability of objects, following standard practices in VOTS benchmarks \cite{dunnhofer2023visual,LaSOT,GOT10k,kristan2015visual,TrackingNet}. These attributes were generated through an automated procedure \cite{dunnhofer2023visual,LaSOT,GOT10k,kristan2015visual,TrackingNet}, and  verified by our team. The selected attributes focus on appearance characteristics influenced by human-object activity videos, which vary depending on the viewpoint. More details in \ref{appendix:attributes} of the Supp. Mat.

\section{Evaluated Methods}
\label{sec:exper}
We evaluated state-of-the-art VOT and VOS methods, ensuring a representation of diverse tracking principles, including those developed for FPV tasks \cite{tang2024egotracks}. We used public code for all trackers with their default hyperparameters to operate each method  under its most optimized and intended settings.
For bounding-box trackers, we selected STARK \cite{stark}, KeepTrack \cite{keeptrack}, the UNICORN box configuration (UNICORN-B) \cite{unicorn}, OSTrack \cite{ostrack}, SeqTrack  \cite{seqtrack}, EgoSTARK \cite{tang2024egotracks}, TaMOs \cite{tamos}, ARTrackV2 \cite{artrackv2}, SAM2 in its box-initialized configuration (SAM2-B) \cite{sam2}, and SAMURAI \cite{samurai}.
For segmentation trackers, we evaluated 
AOT \cite{aot}, the UNICORN mask configuration (UNICORN-M) \cite{unicorn}, XMem \cite{xmem}, Cutie \cite{cutie}, SAM2 in its mask-initialized configuration (SAM2-M) \cite{sam2}, and DAM4SAM \cite{dam4sam}.
To evaluate the impact of viewpoint on training, we selected the STARK \cite{stark} and XMem \cite{xmem} as our baseline methodologies due to their established relevance and adaptability to FPV tracking \cite{tang2024egotracks,dunnhofer2023visual,EgoExo4D}. Trained and finetuned versions of them are referred to as STARK-T-$\pov$, STARK-F-$\pov$, XMem-T-$\pov$, XMem-F-$\pov$.
Further details on these implementations are given in \ref{appendix:methods} of the Supp. Mat.

\section{Results and Discussion}
\label{sec:results}

\paragraph{FPV is challenging for generic object trackers.} Fig. \ref{fig:scatterplotlt} (a)
shows that the values \deltaauc, \deltanps, \deltagsr\ achieved by generic object trackers in typical  human-object activity understanding videos are all negative, indicating that all methods consistently underperform in FPV compared to TPV. A T-Test on $\pov$-AUC scores yielded a $p$-value $< 0.004$, indicating a statistically significant performance difference.
Box and segmentation-based trackers exhibit similar performance drops. Bounding-box trackers show average declines of -9.5, -9.8, -20.7 for \deltaauc, \deltanps, \deltagsr, while segmentation trackers decline by -9.3, -10.5, and -18.7.
Among the metrics, \deltagsr\ exhibits the largest performance decline, indicating that FPV presents significant challenges for maintaining coherent and continuous object tracking.

These results align with previous findings \cite{dunnhofer2023visual,tang2024egotracks,darkhalil2022epic}, confirming that generic object trackers, both bounding-box and segmentation-based, struggle with accurate tracking in egocentric videos due to FPV viewpoint characteristics.

\paragraph{Trackers learn viewpoint biases.}
Since modern methods all use deep learning and achieve lower FPV results despite architectural and optimization variations, it is important to evaluate the impact of viewpoint-specific training data on tracking behavior.
Fig. \ref{fig:scatterplotlt} and Tab. \ref{tab:trainingshort} present such results for the STARK and XMem methods. 
We observe that training in the generic, TPV, and FPV domains impacts performance differently. Training on generic datasets, such as popular VOT and VOS training sets, results in a bias favoring TPV performance, suggesting that these datasets are not aligned with the conditions of erratic camera motion. Training in the TPV domain maintains this bias, likely due to the limited visual variability across consecutive frames, as TPV videos often involve static cameras. Training in the FPV domain biases the model toward FPV-specific characteristics.
Training on both FPV and TPV reduces viewpoint bias but still favors TPV and fails to match view-specific performance.
Fine-tuning on generic datasets is essential for improving generalization across both viewpoints.
Notably, a model optimized on TPV performs worse in long-term TPV tracking (e.g. STARK-T-TPV TPV-AUC 48.0) than a model optimized on FPV does in long-term FPV tracking (e.g. STARK-T-FPV FPV-AUC 49.8). Contrarily to \cite{dunnhofer2023visual,tang2024egotracks,darkhalil2022epic}, this result shows that long-term FPV tracking is not  more challenging than long-term TPV tracking.
In short-term conditions, a TPV-trained tracker outperforms an FPV one in its respective view, indicating that viewpoint training impacts re-detection differently across views.

Overall, a model trained in either the FPV or TPV domain achieves higher performance when tested in the corresponding domain, indicating that the trackers learn biases of the specific viewpoint.

\begin{table}
\caption{\textbf{Training datasets cause viewpoint biases.} We evaluate STARK's and XMem's FPV-AUC, TPV-AUC, $\Delta_{\text{AUC}}$ after being trained on scenes captured by synchronized FPV and TPV.}
    \label{tab:trainingshort}
    \centering
    \fontsize{5}{6}\selectfont
    \setlength\tabcolsep{.1cm}
    \tblalternaterowcolors
    
    \begin{tabular}{c c|c c c | cc c | ccc | ccc }

    \toprule
    
     Training  & Fine-tuning & \multicolumn{6}{c|}{STARK} & \multicolumn{6}{c}{XMem} \\
     \rowcolor{white} Domain & Domain & \multicolumn{3}{c|}{long-term} & \multicolumn{3}{c|}{short-term} & \multicolumn{3}{c|}{long-term} & \multicolumn{3}{c}{short-term}  \\
         
         \midrule

          generic & - & 35.5& 42.8& -7.3 & 56.7& 67.2& -10.5 & 37.1& 45.9& -8.8 & 63.8& 66.8& -3.0 \\ 

         \tpv & - & 36.8& 48.0& -11.2 & 54.4& 67.0& -12.6 & 33.6& 51.5& -17.9 & 58.6&  70.9& -12.3 \\ 

         \fpv & - & 49.8& 38.4& 11.4 & 66.0& 60.8& 5.2 & 56.7& 49.2& 7.5 & 73.6& 70.0& 3.6  \\ 

        \fpv,\tpv & - & 47.9& 50.0& -2.1 & 66.5& 68.7& -2.2 & 54.4& 54.6& -0.2 & 72.9& 72.3& 0.6 \\

        generic  & \tpv & 42.2& 51.7& -9.5 & 59.8& 71.7& -11.9 & 40.2& 52.6& -12.4 & 62.4& 72.0& -9.6 \\ 

        generic  & \fpv & 51.9& 44.2& 7.7 & 70.0& 66.2& 3.8 & 56.3& 52.6& 3.7 & 73.0& 71.8& 1.2  \\ 

        generic  & \fpv,\tpv & 51.4& 51.9& -0.5 & 68.7& 72.4& -3.7 & 54.8& 55.5& -0.7 & 73.0& 73.4& -0.4  \\ 
        
         \bottomrule
   
    \end{tabular}
    
\end{table}

\begin{figure}[t]
    \centering
    \includegraphics[width=1\columnwidth]{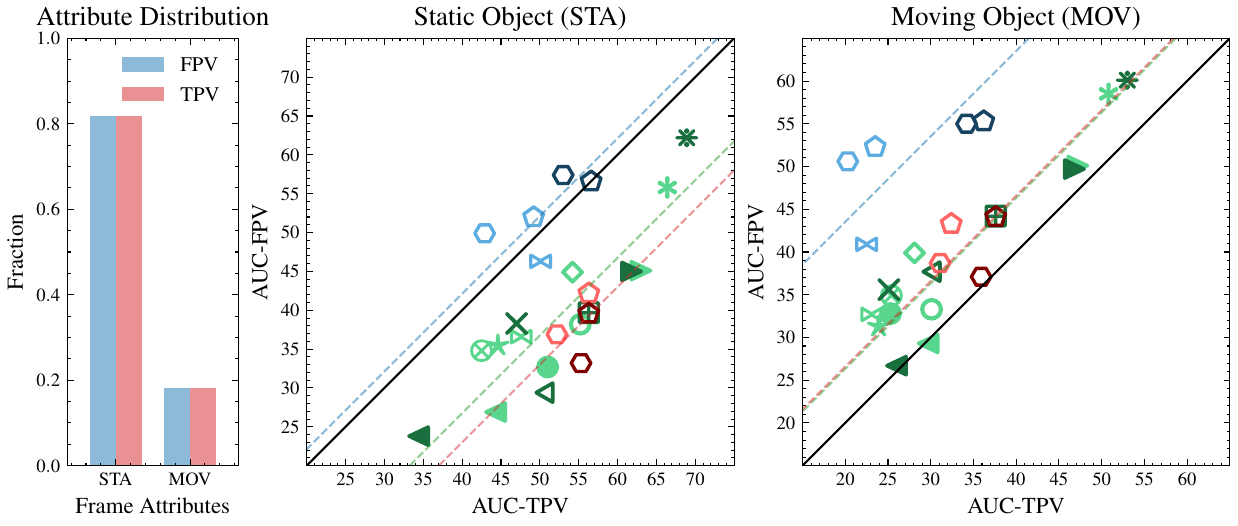}
    \caption{\textbf{FPV is difficult due to the continusouly moving camera, but TPV is not easier when objects move.} We assessed the viewpoint impact on  trackers in frames with objects labeled as being static or moving in the scene. Refer to Fig. \ref{fig:scatterplotlt} for the legend.}
    \label{fig:movsta}
\end{figure}

\begin{figure}[t]
    \centering
    \includegraphics[width=.9\columnwidth]{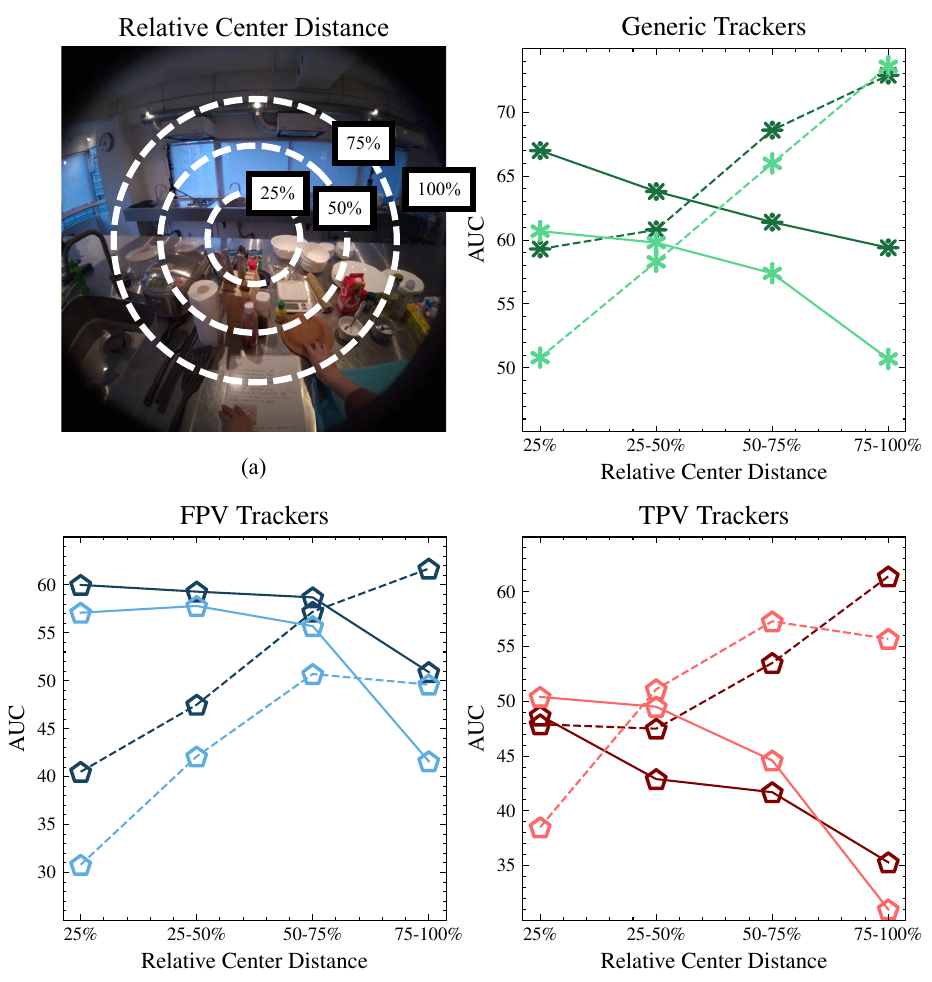}
    \caption{\textbf{The field of view affects tracking in FPV.} We analyzed the impact of object deformation due to distance from the image center on tracking accuracy. (a) shows the distance ranges used to cluster ground-truth mask barycenters. The plots depict AUC as a function of distance for different trackers, with solid lines connecting FPV scores and dashed lines TPV scores. Refer to  Fig. \ref{fig:scatterplotlt} for the symbol and color legend.}
    \label{fig:centerdist}
\end{figure}

\paragraph{Human-object activity videos are challenging, regardless of the viewpoint.}
Comparing the scale of the scores of generic object trackers in Fig. \ref{fig:scatterplotlt} (a) with those from popular benchmarks \cite{LaSOT,xu2018youtube} reveals a significant decline in both FPV and TPV. For example, OSTRack achieves an AUC-FPV of 43.7 and an AUC-TPV of 49.2, down from a 71.1 AUC on the LaSOT benchmark \cite{ostrack}. Similarly, SAM2-M achieves an AUC-FPV of 45.7 and an AUC-TPV of 58.8, lower than its $\mathcal{G}$ score of 89.3 on YouTubeVOS \cite{sam2}. 
This remains true even for the domain-specific trackers in Tab. \ref{tab:trainingshort}.

These results indicate that a significant portion of the performance drop is due challenges of the human-object interaction domain 
rather than only the FPV perspective itself.

\paragraph{FPV tracking is affected by object disappearances.}
Given the challenges introduced by FPV, as shown in Fig. \ref{fig:scatterplotlt} (a), we aim to identify the key factors contributing to this performance difference.
Fig. \ref{fig:scatterplotlt} (b) shows SOPE results on 4,114 short subsequences per viewpoint extracted from the 544 test sequences. Each subsequence consists of consecutive frames where the target remains visible, excluding occlusions and out-of-view periods.
A comparison of the plot scales in Fig. \ref{fig:scatterplotlt} (a) and (b) shows that such events are the main factors contributing to the performance decline on human activity videos, especially in FPV.
When the target is visible, most generic object trackers still show a TPV bias, although to a lesser degree. However, SAM2-based approaches (SAM2, SAMURAI, DAM4SAM) perform slightly better in FPV, indicating minimal viewpoint bias and suggesting that FPV is not more challenging for foundational models when the target is visible. For other methods, the continued TPV bias implies that additional FPV-specific characteristics affect tracking performance.

Overall, the results highlight that addressing object re-detection is crucial to improve tracking accuracy in human-object activity understanding, especially in FPV.

\begin{figure}[t]
    \centering
    \includegraphics[width=\linewidth]{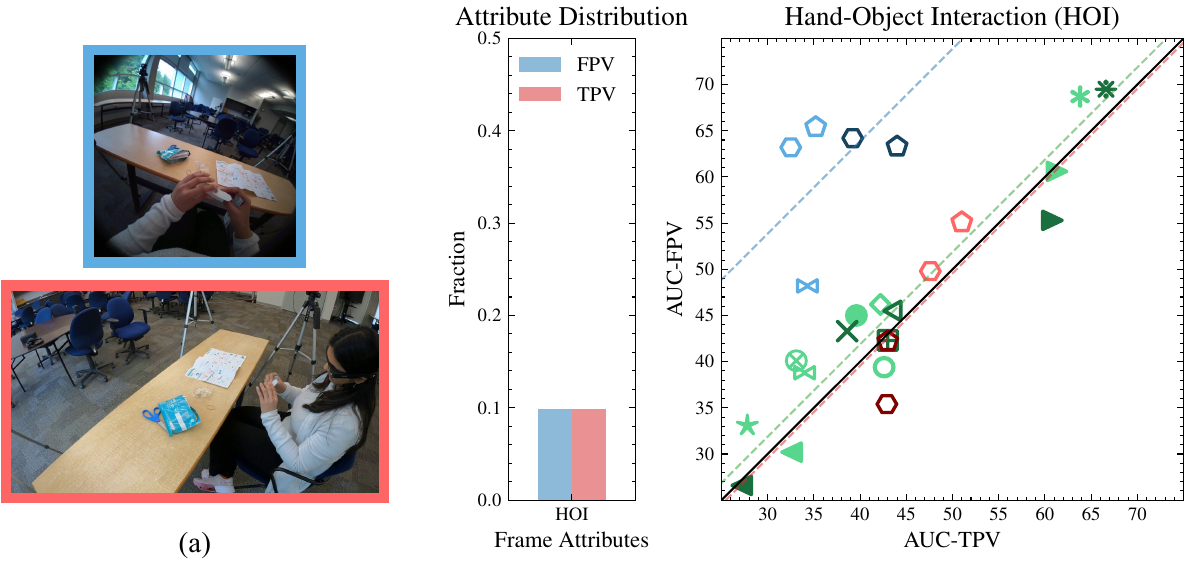}
    \caption{\textbf{Objects involved in hand interactions are more challenging to track in TPV.} We analyzed FPV and TPV tracking performance when the camera wearer interacts with the target object by his/her hands. (a) illustrates the different object appearances when interacted in the two views. Refer to Fig. \ref{fig:scatterplotlt} for the legend.}
    \label{fig:hoi}
\end{figure}

\begin{figure*}[t]
    \centering
    \includegraphics[width=1\linewidth]{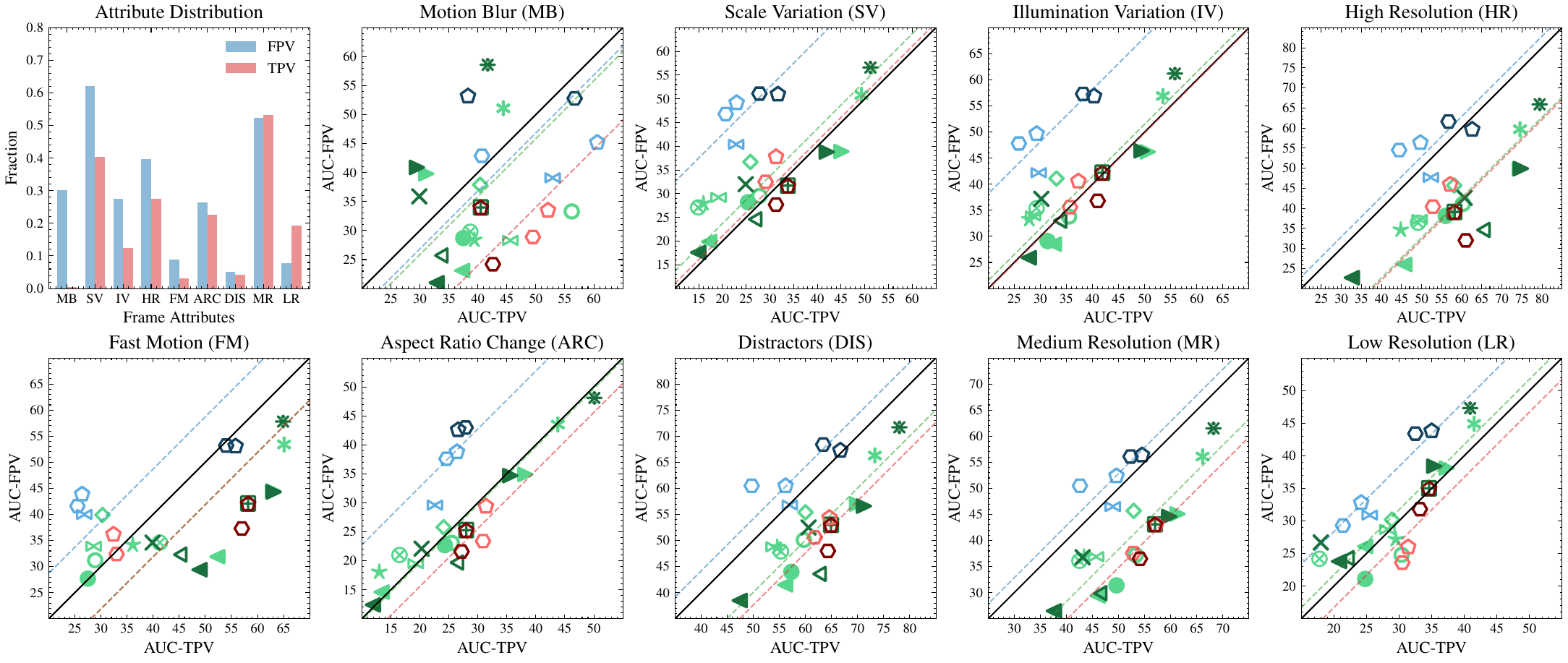}
    \caption{\textbf{The effects of object and camera motion, along with distractors, pose significant challenges in FPV.} We analyzed tracking performance in relation to frame attributes to identify visual variability factors specific of FPV and TPV. Refer to Fig. \ref{fig:scatterplotlt} for the legend.}
    \label{fig:scatterplotattr}
\end{figure*}

\paragraph{FPV involves challenging egomotion, but object displacements are more difficult in TPV.}
Since target objects in human-object activity domains are predominantly rigid \cite{EgoExo4D, grauman2022ego4d, tang2024egotracks, darkhalil2022epic}, one might question whether the performance gap between FPV and TPV arises because objects in TPV typically remain static, making tracking easier. The histogram in Fig. \ref{fig:movsta} supports this hypothesis. When the object is static (STA) in the scene, TPV performance is generally better, indicating that FPV is more challenging due to the dynamic, continuously moving camera that alters the object's appearance. Training in FPV helps addressing this variability. However, when the object is moving (MOV), FPV performance tends to be higher. In this situation, object motion is frequently associated with hand interactions, which tend to keep the object within the camera's field of view, thereby facilitating tracking.

Overall, the continuous motion of the camera is a distinctive characteristic of FPV, and tracking methods must account for it to enhance accuracy.

\paragraph{FPV's field of view affects tracking.}
FPV cameras typically have a wider field of view \cite{aria,EgoExo4D,damen2018scaling} than TPV cameras because they are positioned closer to the action and require a broader perspective to capture sufficient visual information. This setup introduces image distortion, particularly towards the edges of the frame, affecting object appearance. Fig. \ref{fig:centerdist} examines how tracking performance varies in relation to object position in FPV and TPV frames. In FPV, performance declines as objects move away from the center, whereas TPV shows an opposite trend. While training on FPV videos can partially mitigate this effect, objects with significant distortions remain challenging to track. 

Overall, these findings reveal further differences in tracking behavior between FPV and TPV. We demonstrate that FPV tracking is also influenced by the camera setup.

\paragraph{Hand-object interaction are more challenging to track in TPV.}
Hand-object interactions (HOIs) are considered as particular characteristics of egocentric vision \cite{damen2018scaling,grauman2022ego4d,dunnhofer2023visual}, and often considered as challenging situations to overcome because of the hands partially or even completely occluding objects. However, as shown in Fig. \ref{fig:hoi}, tracking objects manipulated by hands is actually easier in FPV than in TPV, likely due to the higher resolution interacted objects have in FPV,  as visualized by the example in Fig. \ref{fig:hoi} (a). FPV-trained trackers exhibit a bias toward these interactions and struggle to generalize to similar situations in TPV.

The tracking of objects involved in hand interactions is not a particular challenge of egocentric FPV.

\paragraph{Object and camera motion, and distractors are challenges in FPV.}
We examined tracking behavior across various visual variability factors to identify those unique to each viewpoint. The results are shown in Fig. \ref{fig:scatterplotattr}. As depicted in the top-left histogram, FPV and TPV produce different object appearance distributions. FPV is characterized by more motion blur (MB), scale variation (SV), high object resolution (HR), fast motion (FM), aspect ratio change (ARC), illumination variation (IV), and the presence of distractors (DIS). Among these factors, MB, HR, FM, and DIS pose greater challenges in FPV, as trackers perform better in TPV under these conditions. In contrast, SV, ARC, and LR have less impact on FPV tracking performance. 

Consistent with previous studies \cite{dunnhofer2023visual}, we confirm that fast motion, motion blur, and the presence of distractors are key factors affecting FPV tracking performance.

\section{Conclusions}
In this paper, we investigated if egocentric FPV is really more challenging for VOTS. Our findings reveal that the answer is not simply ``yes'' or ``no''.
We confirmed a performance decline in state-of-the-art generalistic trackers when evaluated on FPV. One contributing factor is the broader human-object interaction domain, which is insufficiently represented in training datasets.
Foundation model-based trackers are less impacted by the video viewpoint, and show no FPV-induced drop when the object is visible. In contrast, small-scale models show a stronger bias towards TPV.
Training data distributions introduce viewpoint biases, with models performing best in the viewpoint used for optimization. Surprisingly, TPV-optimized trackers face more challenges than FPV-optimized ones.
We identified distinctive challenges in FPV tracking, with object disappearances being a major cause of performance drop. Continuous camera motion alters object appearance, reducing tracking accuracy. Fast appearance changes and distractors also negatively impact FPV tracking, while hand-object interactions are not as problematic than previously stated. Additionally, the camera's wide field of view distorts objects near the frame edges, further affecting FPV tracking accuracy.

\newpage
\noindent\paragraph{Acknowledgments.} This research has been funded by the European Union, NextGenerationEU – PNRR M4 C2 I1.1, RS Micheloni. Progetto PRIN 2022 PNRR - ``Tracking in Egovision for Applied Memory (TEAM)'' Codice P20225MSER\_001 Codice CUP G53D23006680001. 
Matteo Dunnhofer received funding from the European Union’s Horizon Europe research and innovation programme under the Marie Skłodowska-Curie grant agreement n. 101151834 (PRINNEVOT CUP G23C24000910006).

{
    \small
    \bibliographystyle{ieeenat_fullname}
    \bibliography{main}
}

\clearpage
\setcounter{page}{1}
\maketitlesupplementary

\appendix

In this appendix, we present additional motivations, details, and results regarding our benchmark study.

We emphasize that the goal of this paper is not to develop new tracking methods for FPV but to enhance understanding of the task of object tracking in FPV. We aim to provide deeper insights that can better guide the future development of object tracking algorithms.

\section{Details on the \benchmarkname\ Benchmark}
\label{appendix:benchmark}

\subsection{Background} 
Previous VOT and VOS benchmarking studies in egocentric vision have employed the following method to quantify the performance of a tracking algorithm. They considered an annotated video as a pair $\video = (\frames, \annos)$ where 
   $\frames = \{\frame_t\}_{t=0}^{T-1} \text{ and } \annos = \{\anno_t\}_{t=0}^{T-1}$
are, respectively, a sequence of $T$ RGB frames $\frame_t$, and a sequence of $T$ annotations $\anno_t$ in the form of bounding-boxes $\bbox_t$ \cite{dunnhofer2023visual,tang2024egotracks} or segmentation masks $\mask_t$ \cite{darkhalil2022epic}.
Annotations are provided for all or a subset of the frames, in the latter case $\anno_t = \emptyset$ for some $t$.  
The evaluation protocol used first initialized the tracker using the first frame $\frame_0$ and its corresponding annotation $\anno_0$.
Then, the tracker was run on all subsequent frames $\frame_t, t > 0$, producing a set of predictions $\preds =\{\pred_t\}_{t=1}^{T-1}$ represented as boxes $\widehat{\bbox}_t$ or masks $\widehat{\mask}_t$. 
This protocol, adopted from popular VOT and VOS benchmarks, is defined as the one-pass evaluation (OPE) protocol in VOT \cite{wu2013online, wu2015} and as semi-supervised evaluation in VOS \cite{perazzi2016benchmark}.
To obtain a score expressing the quality of the behavior of the algorithm, the predictions were compared to the ground-truth annotations using a scoring function  %
$\sigma\big(\{ \pred_t\}_{t=1}^{T-1}, \{ \anno_t\}_{t=1}^{T-1}\big)$.
This process was repeated across multiple sequences, and the scores were averaged to produce a single metric that quantifies the algorithm's overall performance \cite{dunnhofer2023visual,tang2024egotracks,darkhalil2022epic}. 

The benchmarking efforts employing this schema \cite{dunnhofer2023visual,tang2024egotracks,darkhalil2022epic} compared the obtained scores with those achieved with the same protocol on established VOT and VOS benchmarks \cite{wu2013online,kristan2019seventh,LaSOT,perazzi2016benchmark,xu2018youtube}, highlighting a performance decline used to claim that FPV is more challenging than TPV.
However, this comparison has several limitations. The overall scores come from different data domains, leading to inconsistencies due to differences in object categories and behaviors, sequence lengths, annotation rates, and dataset sizes. Additionally, training sets for training tracking models were drawn from mismatched data distributions. These factors can obscure the true impact of the FPV viewpoint and potentially mislead conclusions about VOTS algorithm performance in egocentric vision.
It is worth mentioning that these issues may affect any benchmark dataset that differs significantly from established ones. In this paper, we specifically focus on egocentric FPV because it was often claimed to be particularly challenging.

\subsection{Single Object Tracking}
\label{appendix:sot}
In this paper, we focus on tracking a single object per video. This choice to restrict the analysis to a single object is to obtain a more detailed examination of the key challenges and factors affecting FPV and TPV tracking. This approach ensures that the evaluation remains unaffected by the complexities introduced by multi-object interactions.
Future work could explore multi-object tracking (MOT) evaluation approaches \cite{luiten2021hota} to achieve a more comprehensive understanding of the impact of FPV and TPV on MOT algorithms. We believe that the insights provided in this study will be valuable for the development an benchmarking of such methods in FPV and TPV.

\subsection{Online Evaluation and Initialization}
In designing SOPE, we adhere to the OPE \cite{wu2015} and semi-supervised protocols \cite{perazzi2016benchmark}, which process video frames sequentially in an online manner. This ensures fair comparison with previous benchmark studies \cite{dunnhofer2023visual,tang2024egotracks,darkhalil2022epic} while also reflecting real-world scenarios where VOTS algorithms must operate in real-time, processing streaming video from wearable cameras for timely video understanding and user assistance.

For tracker initialization in SOPE, we follow again the OPE \cite{wu2015} and semi-supervised protocols \cite{perazzi2016benchmark}, where the target is initialized in the first frame of the sequence. While user-provided initialization is less common in egocentric vision, prior work \cite{dunnhofer2023visual,manigrasso2024online,goletto2024amego} has shown that visual trackers can be initialized by object detectors in the context of tracking-based downstream tasks. Thus, SOPE's standard initialization—using the target's first appearance and initial localization in the two views—remains relevant with respect to the real-world usage of a tracker.

\subsection{Video Collection}
\label{appendix:video}
\begin{figure}[t]
    \centering
    \includegraphics[width=\columnwidth]{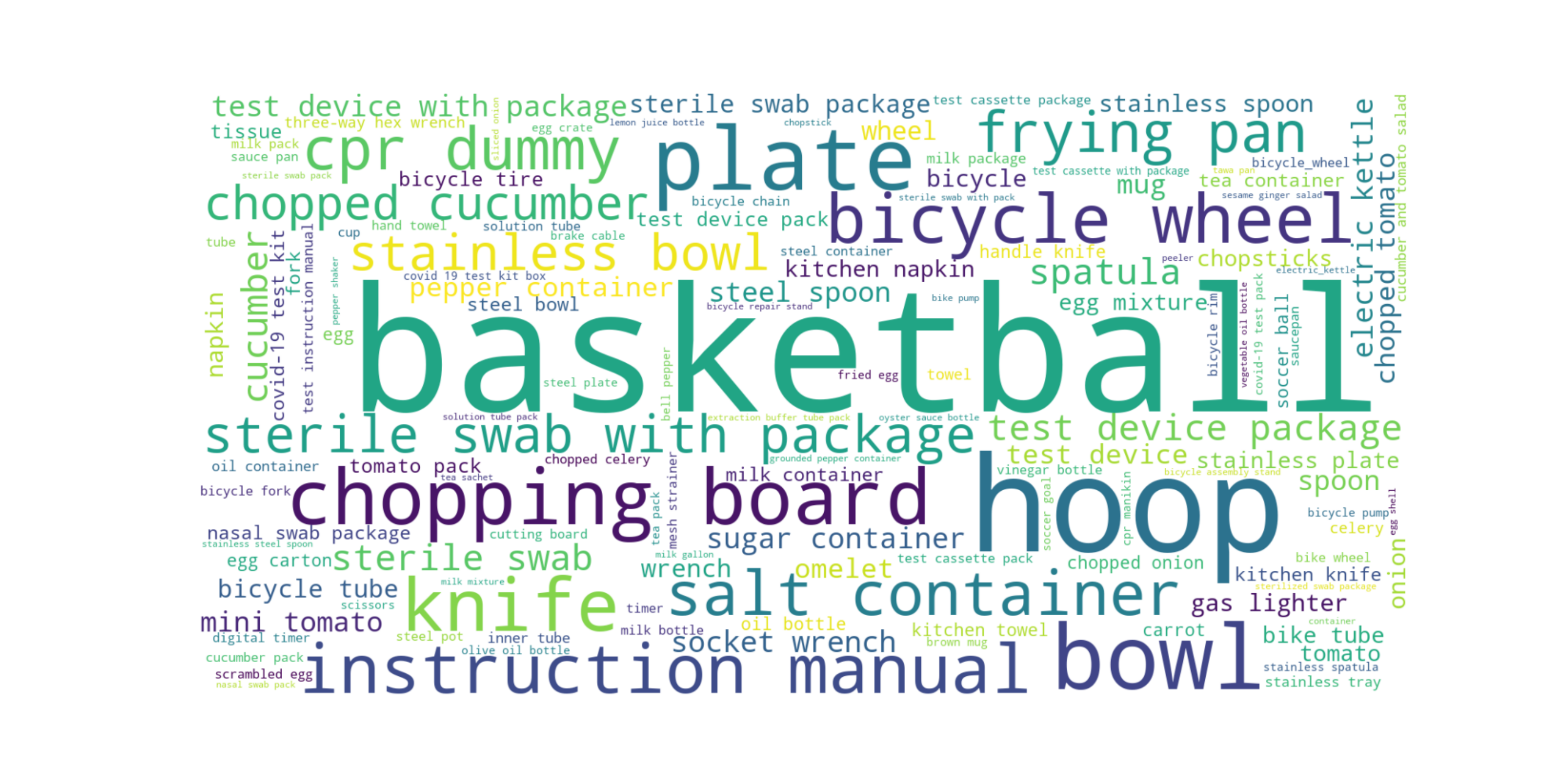}
    \caption{\textbf{Object categories represented in \benchmarkname.} This wordcloud visualizes the categories and the frequency of the target objects available in our benchmarks's training and test sets.}
    \label{fig:wordcloud}
\end{figure}

The video sequences in the \benchmarkname\ benchmark were selected from the EgoExo4D dataset \cite{EgoExo4D}, which is currently the largest resource for studying and developing human activity understanding algorithms from synchronized egocentric (FPV) and exocentric (TPV) point of views.
It contains 1,422 hours of video featuring diverse activities such as sports, music, dance, and bike repair, collected from over 800 participants across 13 cities worldwide. Object categories relate to the activities performed in the videos and include kitchen tools, working tools, appliances, sport equipment, kit parts, etc.
The \benchmarkname\ dataset consists of 6 human-object activities — bike repair, cooking, basketball, cardiopulmonary resuscitation, COVID testing, and soccer — featuring circa 285 distinct object categories (see Fig. \ref{fig:wordcloud}).
The videos are captured by 181 unique camera wearers in 53 different environments across 12 institutions.
Ego-Exo4D's extensive coverage makes it a unique dataset, capturing a wide range of real-world scenarios with different individuals and various object types from both FPV and TPV perspectives. This diversity allowed us to curate a set of FPV and TPV tracking sequences that accurately reflect real-world application scenarios.

The FPV videos in Ego-Exo4D are recorded using Aria smart glasses \cite{aria}, equipped with an 8 MP RGB camera capturing frames at a resolution of $1408\times1408$. For TPV perspectives, multiple stationary GoPro cameras are used, producing landscape videos with a resolution of $1920\times1280$. The placement and number of these exocentric cameras vary per scenario to ensure optimal coverage without obstructing participants' activities \cite{EgoExo4D}. In each FPV-TPV pair, we select the TPV view that was originally annotated with object tracks in Ego-Exo4D, as it provides the clearest observation of the scene \cite{EgoExo4D}. To optimize storage, we performed experiments by resizing FPV frames to $720\times720$ and TPV frames to $1280\times720$.

\subsection{Bounding-box Annotations}
The ground-truth axis-aligned bounding-box annotations $\{\bbox^{\pov}_t \}_{t=0}^{T-1}$ for each FPV-TPV video in \benchmarkname\ are derived by determining the minimum and maximum $x, y$ coordinates of the positive pixels in the corresponding segmentation masks $\{\mask^{\pov}_t \}_{t=0}^{T-1}$.

\subsection{Frame Attributes}
\label{appendix:attributes}
The sequences have been annotated with 12 attributes that characterize motion and visual appearance changes affecting the target object. These attributes help analyze tracker performance under various conditions that may impact its behavior. Each attribute has been assigned on a per-frame basis to enable a more robust evaluation, following the approach in \cite{kristan2016novel}. They have been selected from commonly used attributes in previous tracking benchmarks \cite{wu2015,kristan2016novel,LaSOT,GOT10k} to ensure they capture scenarios relevant to both FPV and TPV videos.
Following \cite{LaSOT,kristan2016novel,GOT10k}, the attributes were initially assigned using an automatic approach and later verified by our research team, consisting of two postdoctoral researchers and a full professor with expertise in visual tracking. Below, we describe the characteristics each attribute represents. The procedures outlined are applied independently to the FPV and TPV sequences.
\begin{itemize}

\item{\textit{Scale Variation (SV).}} A scale variation in the object's appearance occurs if the ratio of the bounding-box area between the first and the current frame falls outside the range [0.5, 2] \cite{wu2015,LaSOT}.

\item{\textit{Aspect Ratio Change (ARC).}} An aspect ratio change occurs if the ratio of the bounding-box aspect ratio between the first and the current frame falls outside the range [0.5, 2] \cite{wu2015,LaSOT}.

\item{\textit{Illumination Variation (IV).}} Illumination variation occurs if the target's bounding-box is subject to significant light changes. The degree of illumination variation in each frame is measured by the change in average color between the first and the current target patch, following \cite{GOT10k}. A threshold of 0.15 is used to determine illumination variation.

\item{\textit{Distractors (DIS).}} A frame contains distractors if it includes objects similar to the target, either from the same category or with a visually similar appearance. To identify distractors, we run a SAM2 instance \cite{sam2} on each frame using a dense grid of point prompts to extract candidate object positions. Each candidate is then verified using DinoV2 \cite{oquab2023dinov2} by computing the cosine similarity between its extracted features and those of the target crop from the first frame. A candidate is considered a distractor if its cosine similarity exceeds 0.5 and its bounding-box overlap with the ground-truth is below 0.5.

\item{\textit{Motion Blur (MB).}} Motion blur occurs when the target region appears blurred due to object or camera motion. Following \cite{kristan2015visual,kristan2016novel}, we detect motion blur by computing the variance of the Laplacian on the target patch in the current frame. A threshold of 100 is used to determine the presence of motion blur.

\item{\textit{Fast Motion (FM).}} Fast motion is detected when the target bounding-box moves a distance greater than its own size between consecutive frames \cite{LaSOT}. This attribute is computed by measuring the displacement of the bounding box during periods of target visibility.

\item{\textit{Low Resolution (LR).}} The target patch is considered low resolution if the area of the target bounding-box is smaller than $32^2$ pixels \cite{coco}.

\item{\textit{Medium Resolution (MR).}} The target patch is considered medium resolution if the area of the target bounding-box is between $32^2$ and $96^2$ pixels \cite{coco}.

\item{\textit{High Resolution (HR).}} The target patch is considered high resolution if the area of the target bounding-box is larger than $96^2$ pixels \cite{coco}.

\end{itemize}

\begin{table*}
\caption{\textbf{Effect of score weighting by sequence length.} This table shows how performance difference scores change when each sequence score is weighted by its annotation length. We apply this weighting because trackers behave differently across the two viewpoints, and unweighted scores from short or long videos can distort the true average performance.} 
    \label{tab:weighting}
    \centering
    
    \tblalternaterowcolors

    \setlength\tabcolsep{.3cm}
    
    \begin{tabular}{c l| c |ccc|ccc|ccc}

    \toprule
    
    & & & \multicolumn{3}{c|}{AUC} & \multicolumn{3}{c|}{NPS} & \multicolumn{3}{c}{GSR} \\
         \rowcolor{white} & \multirow{-2}{*}{Tracker} & \multirow{-2}{*}{Weight} & FPV & TPV & $\Delta_{\text{AUC}}$ & FPV & TPV & $\Delta_{\text{NPS}}$ & FPV & TPV & $\Delta_{\text{GSR}}$ \\
         
         \midrule

 & & \cmark & 51.9 & 44.2 & 7.7 & 54.0 & 45.6 & 8.4 & 11.3 & 28.7 & -17.4 \\
 \cellcolor{white} \multirow{-2}{*}{\textcolor[HTML]{5dade2}{$\pmb{\pentagon}$}} & \cellcolor{white}  \multirow{-2}{*}{STARK-F-FPV} & \xmark & 54.0 & 43.7 & 10.3 & 56.0 & 45.1 & 10.9 & 21.3 & 33.7 & -12.4 \\ 

 \midrule

  & & \cmark & 56.3 & 52.6 & 3.7 & 60.8 & 57.6 & 3.2 & 29.8 & 39.4 & -9.6 \\ 
 \cellcolor{white}  \multirow{-2}{*}{\textcolor[HTML]{154360}{$\pmb{\pentagon}$}} & \cellcolor{white}  \multirow{-2}{*}{XMem-F-FPV} & \xmark & 58.7 & 51.9 & 6.8 & 63.6 & 57.0 & 6.6 & 42.9 & 45.6 & -2.7 \\ 

\midrule

  & & \cmark & 61.7 & 65.8 & -4.1 & 66.2 & 71.1 & -4.9 & 35.2 & 52.9 & -17.7 \\ 
 \cellcolor{white} \multirow{-2}{*}{\textcolor[HTML]{196f3d}{$\pmb{\divideontimes}$}} & \cellcolor{white} \multirow{-2}{*}{DAM4SAM} & \xmark & 64.4 & 63.5 & 0.9 & 69.6 & 69.0 & 0.6 & 47.3 & 57.6 & -10.3 \\ 

\midrule

 & & \cmark & 56.2 & 63.3 & -7.1 & 58.3 & 65.6 & -7.3 & 31.8 & 48.2 & -16.4 \\ 

   \cellcolor{white}   \multirow{-2}{*}{\textcolor[HTML]{58d68d}{$\pmb{\ast}$}} & \cellcolor{white}  \multirow{-2}{*}{SAMURAI} & \xmark & 60.6 & 63.1 & -2.5 & 63.1 & 65.3 & -2.2 & 43.9 & 53.5 & -9.6 \\

\midrule

  & & \cmark & 42.2 & 51.7 & -9.5 & 43.3 & 53.9 & -10.6 & 7.9 & 33.0 & -25.1 \\

 \cellcolor{white} \multirow{-2}{*}{\textcolor[HTML]{ff6666}{$\pmb{\pentagon}$}} & \cellcolor{white} \multirow{-2}{*}{STARK-F-TPV} &  \xmark & 44.5 & 50.6 & -6.1 & 45.8 & 52.4 & -6.6 & 16.7 & 37.7 & -21.0 \\

 \midrule

 & & \cmark & 40.2 & 52.6 & -12.4 & 45.7 & 58.5 & -12.8 & 18.9 & 39.7 & -20.8 \\ 

  \cellcolor{white} \multirow{-2}{*}{\textcolor[HTML]{800000}{$\pmb{\pentagon}$}} & \cellcolor{white} \multirow{-2}{*}{XMem-F-TPV} & \xmark & 45.0 & 53.0 & -8.0 & 50.8 & 59.0 & -8.2 & 30.1 & 46.4 & -16.3 \\

         \bottomrule
   
    \end{tabular}
    
\end{table*}

In addition to these standard attributes, we computed the following:

\begin{itemize}

\item{\textit{Static Object (STA) and Moving Object (MOV).}} The object is considered static in the current frame if it remains in the same position relative to the previous frame. This is determined by computing the IoU between the bounding-box of the current and previous frame. If the IoU is above 0.5, the target object is labeled as static; otherwise, it is labeled as moving. This information is computed using only the TPV view, leveraging the synchronization of frames and annotations to assign the corresponding label to the associated FPV frame. Since the TPV camera is stationary, any change in annotation overlap during target visibility periods is due to the motion of the object.

\item{\textit{Hand-Object Interaction (HOI).}} The target object is considered to be in interaction with a person's hands. Following \cite{dunnhofer2023visual,goletto2024amego}, we compute this attribute by first running a hand-object interaction detector \cite{shan2020understanding}. To determine whether the target object is being interacted with, we check if its bounding-box overlaps with the detected object bounding boxes by more than 0.5 IoU and if an interaction state is detected for at least two consecutive annotations (equivalent to a period of 1 second). Once an interaction label is assigned, it remains active for all subsequent frames until two consecutive overlaps fall below 0.5 with no interaction state detected \cite{goletto2024amego}.
This information is computed using only the FPV view becuase the egocentric viewpoint enables a closer view of the interaction between hands and objects \cite{shan2020understanding}. We leverage the synchronization of frames and annotations to assign the corresponding label to the associated TPV frame.
\end{itemize}

\subsection{Metrics}
\label{appendix:metrics}
As shown in Eq. \ref{eq:diff}, we compute the mean signed difference for each sequence, weight it by the annotation length, and then average it across the sum of all the annotation lengths \cite{kristan2016novel}.
We applied this weighting because we observed that methods tend to perform better on short FPV videos. Without weighting, a high score from a short FPV sequence carries the same influence as a low score from a long FPV sequence, which can mask the overall lower tracking accuracy of longer videos. 
Conversely, algorithms tend to perform better on long TPV videos. In this case, a long TPV sequence with limited object motion and a high score would carry the same weight as a short TPV sequence with object motion and poor performance, potentially overshadowing the true performance trend.
In Tab. \ref{tab:weighting}, we report the difference in using or not using the weighting on the AUC, NPS, and GSR metrics. Tab. \ref{tab:davis} shows the impact of not having the weighting as used by the standard VOS benchmark evaluation \cite{perazzi2016benchmark,xu2018youtube}. 

The same weighting strategy described before is applied to the standard metrics reported in all tables and figures of this paper, specifically AUC-${\pov}$, NPS-${\pov}$, or GSR-${\pov}$. This is represented by the following equation:
\begin{align}
s^{\pov}_{\sigma} = \frac{1} {\omega_0 + \cdots + \omega_{N-1}}\sum_{i=0}^{N-1} s_{\sigma,i}^{\pov} \cdot \omega_i, \omega_i = |\annos_i^{pov}|
\end{align}  
where $s_{\sigma,i}^{\pov}$ represents the AUC, NPS, or GSR score for an individual sequence.

We compute the sequence-wise scores $s_{\sigma,i}^{\pov}$ for bounding-box trackers by calculating the AUC, NPS, and GSR based on the IoU between the predicted bounding-boxes $\{\widehat{\bbox}^{\pov}_t\}_{t=1}^{T-1}$ and the ground-truth bounding-boxes $\{\bbox^{\pov}_t\}_{t=1}^{T-1}$. For trackers that output segmentation masks, we compute the AUC, NPS, and GSR based on the IoU between the predicted segmentation masks $\{\widehat{\mask}^{\pov}_t\}_{t=1}^{T-1}$ and the ground-truth masks $\{\mask^{\pov}_t\}_{t=1}^{T-1}$. This approach ensures a fair evaluation of the tracker's predictions by comparing them to the ground-truth target state representation that the tracker was optimized for.

In addition to the previously mentioned metric, Table \ref{tab:davis} reports the scores $\mathcal{J} \& \mathcal{F}$, $\mathcal{J}$, and $\mathcal{F}$, which are commonly used for VOS evaluation. These scores were computed as originally described in \cite{perazzi2016benchmark, xu2018youtube}. As with the other metrics, we calculate the mean signed differences $\Delta_{\mathcal{J} \& \mathcal{F}}, \Delta_{\mathcal{J}}, \Delta_{\mathcal{F}}$ to quantify the performance differences between FPV and TPV based on these metrics. For these segmentation-oriented metrics, we convert the bounding-boxes predicted by box-based trackers into segmentation masks by filling the rectangular area within the bounding-box with positive pixels \cite{kristan2020eighth,kristan2023first}.

\section{Details on the Evaluated Methods}
\label{appendix:methods}
For SAM2-based instances \cite{sam2} (SAM2-B, SAM2-M, SAMURAI, DAM4SAM), we used the SAM 2++ Hiera Large instance.
SAM2-B is a variant of SAM2 initialized with a bounding box and producing bounding-box outputs. For SAMURAI, we follow the running setup from the original paper \cite{samurai}, initializing it with a bounding-box and retrieving output bounding-boxes from it.

In the following, we provide  details about the viewpoint-optimized tracking baselines mentioned in Sec. \ref{sec:exper} of the main paper.
For STARK-T-FPV, we train the STARK-ST50 instance \cite{stark} starting from random weights on the FPV sequences of the $\trainset$ in the \benchmarkname\ benchmark. The training consists of 100 epochs in stage 1 and 10 epochs in stage 2. Apart from the number of epochs, all other hyperparameters are kept fixed as originally proposed \cite{stark}.
For STARK-T-TPV, we use the same training configuration, with the only change being the training set, which consists of the TPV sequences from  $\trainset$. 
For STARK-T-FPV,TPV (row 4 of Tab. \ref{tab:trainingshort}), we use the same training configuration as described previously, with the only change being the training set, which includes both FPV and TPV sequences from $\trainset$.
For STARK-F-FPV, STARK-F-TPV, and STARK-F-FPV,TPV, we follow the same approach as before, but start with pretrained model weights obtained after training for generic object tracking on the TrackingNet \cite{TrackingNet}, LaSOT \cite{LaSOT}, GOT-10k \cite{GOT10k}, and COCO \cite{coco} datasets.
The original code repository was used to implement all of these procedures.\footnote{\url{https://github.com/researchmm/Stark}}

For the XMem baseline, we follow a similar approach.
For XMem-T-FPV, we the ResNet50-based instance \cite{xmem} starting from weights pretrained on static images (stage 0) on the FPV sequences of  $\trainset$. The training kept all hyperparameters fixed as originally proposed \cite{xmem}.
For XMem-T-TPV, we use the same training configuration, with the only change being the training set, which consists of the TPV sequences from  $\trainset$. 
For XMem-T-FPV,TPV (row of Tab. \ref{tab:trainingshort}), we use the same training configuration as described previously, with the only change being the training set, which includes both FPV and TPV sequences from $\trainset$.
For XMem-F-FPV, XMem-F-TPV, and XMem-F-FPV,TPV, we follow the same approach as before, but start with pretrained model weights obtained after training on generic object VOS using the DAVIS \cite{perazzi2016benchmark}, and YouTube-VOS \cite{xu2018youtube} datasets (stage 3).
The original code repository was used to implement all of these procedures.\footnote{\url{https://github.com/hkchengrex/XMem}}

All the code used for this study was implemented in Python and run on a machine with an Intel Xeon E5-2690 v4 @ 2.60 GHz CPU, 320 GB of RAM, and 6 NVIDIA TITAN V GPUs.

\section{Details and Additional Results}
\label{appendix:results}

In all Figures and Tables in Sec. \ref{sec:results} of the main paper, unless stated otherwise, results are based on the 544 pairs in $\testset$ under the long-term object tracking setting.

\paragraph{Frame-based attribute evaluation.}
For experiments involving frame-based attributes, scores are computed using only frames labeled with the respective attribute. In these cases, each sequence is weighted based on the number of annotated frames containing the attribute of interest.

\paragraph{Long-term tracking scores.} For improved readability, the full version of the scores shown in brackets in Fig. \ref{fig:scatterplotlt} is provided in Tab. \ref{tab:generictrackerlt}. Refer to Sec. \ref{sec:results} of the main paper for a detailed discussion.

\begin{table*}
\caption{\textbf{Long-term object tracking performance across FPV and TPV.} For a better readibility, this table reports the score presented in Figure \ref{fig:scatterplotlt} (a). \textcolor[HTML]{5dade2}{Light blue} represents FPV bounding-box trackers; \textcolor[HTML]{154360}{dark blue} represents FPV segmentation trackers; \textcolor[HTML]{ff6666}{light red} represents TPV bounding-box trackers; \textcolor[HTML]{800000}{dark red} represents TPV segmentation trackers; \textcolor[HTML]{58d68d}{light green} represents generic bounding-box trackers; \textcolor[HTML]{196f3d}{dark green} represents generic segmentation trackers. Trackers are ordered in descending order by $\Delta_{\text{AUC}}.$} 
    \label{tab:generictrackerlt}
    \centering
    
    \tblalternaterowcolors

    \setlength\tabcolsep{.35cm}
    
    \begin{tabular}{c l|ccc|ccc|ccc}

    \toprule
    
    & & \multicolumn{3}{c|}{AUC} & \multicolumn{3}{c|}{NPS} & \multicolumn{3}{c}{GSR} \\
         \rowcolor{white} & \multirow{-2}{*}{Tracker} & FPV & TPV & $\Delta_{\text{AUC}}$ & FPV & TPV & $\Delta_{\text{NPS}}$ & FPV & TPV & $\Delta_{\text{GSR}}$ \\
         
         \midrule

       \textcolor[HTML]{5dade2}{$\pmb{\hexagon}$} & STARK-T-FPV & 49.8 & 38.4 & 11.4 & 51.8 & 39.4 & 12.4 & 10.9 & 24.8 & -13.9 \\ 

 \textcolor[HTML]{5dade2}{$\pmb{\pentagon}$} & STARK-F-FPV & 51.9 & 44.2 & 7.7 & 54.0 & 45.6 & 8.4 & 11.3 & 28.7 & -17.4 \\ 

 \textcolor[HTML]{154360}{$\pmb{\hexagon}$} & XMem-T-FPV & 56.7 & 49.2 & 7.5 & 61.0 & 54.5 & 6.5 & 30.9 & 37.7 & -6.8 \\ 

 \textcolor[HTML]{154360}{$\pmb{\pentagon}$} & XMem-F-FPV & 56.3 & 52.6 & 3.7 & 60.8 & 57.6 & 3.2 & 29.8 & 39.4 & -9.6 \\ 

 \textcolor[HTML]{5dade2}{$\pmb{\bowtie}$} & EgoSTARK & 45.0 & 44.7 & 0.3 & 46.4 & 46.5 & -0.1 & 10.0 & 29.7 & -19.7 \\ 

 \textcolor[HTML]{196f3d}{$\pmb{\divideontimes}$} & DAM4SAM & 61.7 & 65.8 & -4.1 & 66.2 & 71.1 & -4.9 & 35.2 & 52.9 & -17.7 \\ 

 \textcolor[HTML]{58d68d}{$\pmb{\otimes}$} & TaMOs & 34.5 & 39.0 & -4.5 & 36.2 & 39.5 & -3.3 & 10.6 & 26.8 & -16.2 \\ 

 \textcolor[HTML]{196f3d}{$\pmb{\times}$} & AOT & 37.5 & 42.6 & -5.1 & 38.7 & 43.4 & -4.7 & 11.4 & 30.7 & -19.3 \\ 

 \textcolor[HTML]{58d68d}{$\pmb{\diamond}$} & OSTrack & 43.7 & 49.2 & -5.5 & 45.3 & 51.0 & -5.7 & 10.2 & 31.3 & -21.1 \\ 

 \textcolor[HTML]{58d68d}{$\pmb{\star}$} & KeepTrack & 34.5 & 40.5 & -6.0 & 36.9 & 42.8 & -5.9 & 8.5 & 29.0 & -20.5 \\ 

 \textcolor[HTML]{58d68d}{$\pmb{\ast}$} & SAMURAI & 56.2 & 63.3 & -7.1 & 58.3 & 65.6 & -7.3 & 31.8 & 48.2 & -16.4 \\ 

 \textcolor[HTML]{58d68d}{$\pmb{\bowtie}$} & STARK & 35.5 & 42.8 & -7.3 & 36.6 & 44.6 & -8.0 & 8.1 & 28.1 & -20.0 \\ 

 \textcolor[HTML]{196f3d}{$\pmb{\blacktriangleleft}$} & UNICORN-M & 23.9 & 32.5 & -8.6 & 28.4 & 40.4 & -12.0 & 5.8 & 19.2 & -13.4 \\ 

 \textcolor[HTML]{196f3d}{$\pmb{\boxplus}$} & XMem & 37.1 & 45.9 & -8.8 & 40.3 & 49.4 & -9.1 & 17.8 & 35.2 & -17.4 \\ 

 \textcolor[HTML]{ff6666}{$\pmb{\pentagon}$} & STARK-F-TPV & 42.2 & 51.7 & -9.5 & 43.3 & 53.9 & -10.6 & 7.9 & 33.0 & -25.1 \\ 

 \textcolor[HTML]{ff6666}{$\pmb{\hexagon}$} & STARK-T-TPV & 36.8 & 48.0 & -11.2 & 38.0 & 50.0 & -12.0 & 6.9 & 29.9 & -23.0 \\ 

 \textcolor[HTML]{800000}{$\pmb{\pentagon}$} & XMem-F-TPV & 40.2 & 52.6 & -12.4 & 45.7 & 58.5 & -12.8 & 18.9 & 39.7 & -20.8 \\ 

 \textcolor[HTML]{196f3d}{$\pmb{\blacktriangleright}$} & SAM2-M & 45.7 & 58.8 & -13.1 & 49.1 & 64.1 & -15.0 & 24.0 & 47.9 & -23.9 \\ 

 \textcolor[HTML]{58d68d}{$\pmb{\circ}$} & SeqTrack & 36.9 & 50.3 & -13.4 & 39.0 & 52.8 & -13.8 & 8.1 & 33.7 & -25.6 \\ 

 \textcolor[HTML]{58d68d}{$\pmb{\bullet}$} & ARTrackV2 & 32.4 & 46.0 & -13.6 & 32.6 & 47.7 & -15.1 & 7.5 & 33.9 & -26.4 \\ 

 \textcolor[HTML]{58d68d}{$\pmb{\blacktriangleright}$} & SAM2-B & 45.8 & 59.9 & -14.1 & 47.9 & 62.8 & -14.9 & 22.7 & 43.3 & -20.6 \\ 

 \textcolor[HTML]{58d68d}{$\pmb{\blacktriangleleft}$} & UNICORN-B & 27.0 & 41.3 & -14.3 & 28.1 & 42.1 & -14.0 & 6.7 & 26.1 & -19.4 \\ 

 \textcolor[HTML]{196f3d}{$\pmb{\triangleleft}$} & Cutie & 30.6 & 46.6 & -16.0 & 33.4 & 50.9 & -17.5 & 15.6 & 36.4 & -20.8 \\ 

 \textcolor[HTML]{800000}{$\pmb{\hexagon}$} & XMem-T-TPV & 33.6 & 51.5 & -17.9 & 37.8 & 56.9 & -19.1 & 14.6 & 39.6 & -25.0 \\

         \bottomrule
   
    \end{tabular}
    
\end{table*}

\paragraph{Details on experiments on the field's of view impact.} 
To generate the results shown in Fig. \ref{fig:centerdist} of the main paper, we categorized each annotation based on its distance from the frame center into four regions: (1) within 25\% of the frame width from the center, (2) between 25\% and 50\% of the frame width, (3) between 50\% and 75\% of the frame width, and (4) beyond 75\% of the frame width. The position of each annotation was determined using the coordinates of its barycenter. This clustering process was applied separately to FPV and TPV. To compute the scores for each cluster, we followed the same procedure used for frame attribute-based evaluation.
For FPV, each cluster contains 11\% (25\%), 33\% (25-50\%), 30\% (50-75\%), 26\% (75-100\%) of the total annotated frames. For TPV, each cluster contains 14\% (25\%), 31\% (25-50\%), 26\% (50-75\%), 29\% (75-100\%) of the total annotated frames.

\begin{table*}
\caption{\textbf{Object tracking performance across FPV and TPV with standard VOS metrics.} This Table reports $\mathcal{J}$ \& $\mathcal{F}$, $\mathcal{J}$, and $\mathcal{F}$ generally used in VOS evaluation \cite{perazzi2016benchmark,xu2018youtube}. Similar conclusions to what reported for Fig. \ref{fig:scatterplotlt} can be made for these results. The computation of these metrics does not take into account the length of the sequence, and this can overshadow the real average FPV and TPV performance. \textcolor[HTML]{5dade2}{Light blue} represents FPV bounding-box trackers; \textcolor[HTML]{154360}{dark blue} represents FPV segmentation trackers; \textcolor[HTML]{ff6666}{light red} represents TPV bounding-box trackers; \textcolor[HTML]{800000}{dark red} represents TPV segmentation trackers; \textcolor[HTML]{58d68d}{light green} represents generic bounding-box trackers; \textcolor[HTML]{196f3d}{dark green} represents generic segmentation trackers. Trackers are ordered in descending order by $\Delta_{\mathcal{J} \& \mathcal{F}}.$}
    \label{tab:davis}
    \centering
    
    \tblalternaterowcolors

    \setlength\tabcolsep{.35cm}
    
    \begin{tabular}{c l|ccc|ccc|ccc}

    \toprule
    
    & & \multicolumn{3}{c|}{$\mathcal{J}$ \& $\mathcal{F}$ } & \multicolumn{3}{c|}{$\mathcal{J}$} & \multicolumn{3}{c}{$\mathcal{F}$} \\
         \rowcolor{white} & \multirow{-2}{*}{Tracker} & FPV & TPV & $\Delta_{\mathcal{J} \& \mathcal{F}}$ & FPV & TPV & $\Delta_{\mathcal{J}}$ & FPV & TPV & $\Delta_{\mathcal{F}}$ \\
         
         \midrule

        \textcolor[HTML]{154360}{$\pmb{\hexagon}$} & XMem-T-FPV & 64.2 & 57.3 & 6.9 & 59.0 & 49.3 & 9.7 & 69.4 & 65.2 & 4.2 \\ 

 \textcolor[HTML]{154360}{$\pmb{\pentagon}$} & XMem-F-FPV & 63.9 & 59.3 & 4.6 & 58.7 & 51.2 & 7.5 & 69.1 & 67.4 & 1.7 \\ 

 \textcolor[HTML]{5dade2}{$\pmb{\hexagon}$} & STARK-T-FPV & 35.8 & 31.5 & 4.3 & 34.3 & 25.2 & 9.1 & 37.3 & 37.7 & -0.4 \\ 

 \textcolor[HTML]{5dade2}{$\pmb{\pentagon}$} & STARK-F-FPV & 37.0 & 34.8 & 2.2 & 35.4 & 27.7 & 7.7 & 38.7 & 41.9 & -3.2 \\ 

 \textcolor[HTML]{196f3d}{$\pmb{\divideontimes}$} & DAM4SAM & 69.4 & 70.1 & -0.7 & 64.8 & 63.5 & 1.3 & 74.0 & 76.8 & -2.8 \\ 

 \textcolor[HTML]{5dade2}{$\pmb{\bowtie}$} & EgoSTARK & 31.4 & 33.9 & -2.5 & 30.2 & 27.4 & 2.8 & 32.6 & 40.5 & -7.9 \\ 

 \textcolor[HTML]{196f3d}{$\pmb{\times}$} & AOT & 42.9 & 46.9 & -4.0 & 38.8 & 39.6 & -0.8 & 47.1 & 54.2 & -7.1 \\ 

 \textcolor[HTML]{58d68d}{$\pmb{\otimes}$} & TaMOs & 29.7 & 34.2 & -4.5 & 27.8 & 25.8 & 2.0 & 31.6 & 42.7 & -11.1 \\ 

 \textcolor[HTML]{58d68d}{$\pmb{\ast}$} & SAMURAI & 63.8 & 69.0 & -5.2 & 59.2 & 60.9 & -1.7 & 68.4 & 77.1 & -8.7 \\ 

 \textcolor[HTML]{58d68d}{$\pmb{\diamond}$} & OSTrack & 31.9 & 37.1 & -5.2 & 30.6 & 30.0 & 0.6 & 33.2 & 44.2 & -11.0 \\ 

 \textcolor[HTML]{196f3d}{$\pmb{\boxplus}$} & XMem & 43.4 & 49.5 & -6.1 & 40.1 & 43.6 & -3.5 & 46.7 & 55.4 & -8.7 \\ 

 \textcolor[HTML]{58d68d}{$\pmb{\bowtie}$} & STARK & 27.7 & 34.1 & -6.4 & 26.4 & 27.0 & -0.6 & 29.0 & 41.2 & -12.2 \\ 

 \textcolor[HTML]{196f3d}{$\pmb{\blacktriangleleft}$} & UNICORN-M & 32.8 & 41.0 & -8.2 & 28.1 & 32.7 & -4.6 & 37.6 & 49.2 & -11.6 \\ 

 \textcolor[HTML]{58d68d}{$\pmb{\star}$} & KeepTrack & 28.2 & 36.6 & -8.4 & 26.8 & 29.2 & -2.4 & 29.7 & 44.1 & -14.4 \\ 

 \textcolor[HTML]{ff6666}{$\pmb{\pentagon}$} & STARK-F-TPV & 30.8 & 40.2 & -9.4 & 29.5 & 31.8 & -2.3 & 32.2 & 48.7 & -16.5 \\ 

 \textcolor[HTML]{58d68d}{$\pmb{\blacktriangleleft}$} & UNICORN-B & 25.5 & 35.1 & -9.6 & 23.4 & 27.7 & -4.3 & 27.7 & 42.5 & -14.8 \\ 

 \textcolor[HTML]{196f3d}{$\pmb{\triangleleft}$} & Cutie & 41.0 & 50.6 & -9.6 & 37.8 & 44.6 & -6.8 & 44.3 & 56.6 & -12.3 \\ 

 \textcolor[HTML]{58d68d}{$\pmb{\circ}$} & SeqTrack & 27.1 & 37.4 & -10.3 & 25.5 & 29.6 & -4.1 & 28.6 & 45.2 & -16.6 \\ 

 \textcolor[HTML]{196f3d}{$\pmb{\blacktriangleright}$} & SAM2-M & 54.9 & 65.3 & -10.4 & 51.4 & 58.5 & -7.1 & 58.4 & 72.1 & -13.7 \\ 

 \textcolor[HTML]{ff6666}{$\pmb{\hexagon}$} & STARK-T-TPV & 28.1 & 38.8 & -10.7 & 26.8 & 30.7 & -3.9 & 29.4 & 46.9 & -17.5 \\ 

 \textcolor[HTML]{58d68d}{$\pmb{\bullet}$} & ARTrackV2 & 23.8 & 35.0 & -11.2 & 23.1 & 28.2 & -5.1 & 24.5 & 41.9 & -17.4 \\ 

 \textcolor[HTML]{800000}{$\pmb{\pentagon}$} & XMem-F-TPV & 49.5 & 60.7 & -11.2 & 44.2 & 52.4 & -8.2 & 54.7 & 69.0 & -14.3 \\ 

 \textcolor[HTML]{58d68d}{$\pmb{\blacktriangleright}$} & SAM2-B & 36.9 & 51.1 & -14.2 & 34.9 & 41.9 & -7.0 & 38.8 & 60.3 & -21.5 \\ 

 \textcolor[HTML]{800000}{$\pmb{\hexagon}$} & XMem-T-TPV & 44.0 & 59.3 & -15.3 & 38.7 & 51.5 & -12.8 & 49.3 & 67.1 & -17.8 \\

         \bottomrule
   
    \end{tabular}
    
\end{table*}

\paragraph{VOS-based evaluation results.}
Tab. \ref{tab:davis} presents the performance scores of the selected trackers using the standard semi-supervised evaluation protocol, measured with the $\mathcal{J} \& \mathcal{F}$, $\mathcal{J}$, and $\mathcal{F}$ metrics \cite{perazzi2016benchmark}. The $\mathcal{J}$ metric quantifies the average overlap between the predicted and ground-truth segmentation masks (similar to AUC), while $\mathcal{F}$ assesses the quality of segmentation boundaries. The $\mathcal{J} \& \mathcal{F}$ metric is the average of the two.
To analyze viewpoint-dependent performance differences, we compute the signed difference for these metrics. 

The results confirm the conclusions drawn from Fig. \ref{fig:scatterplotlt} and Tab. \ref{tab:generictrackerlt}. Generic object trackers exhibit a more significant performance drop in FPV compared to TPV. Additionally, the bias introduced by viewpoint-optimized trackers is reflected also in these metrics.
It is important to note that standard VOS evaluation does not weight sequence scores by sequence or annotation length. In this approach, all sequences contribute equally, regardless of their duration or annotation frequency, even though these factor can influence tracking performance scoring. As a result, this evaluation method fails to accurately capture viewpoint bias, as performance measurements become skewed toward short, high-scoring sequences, masking the true impact of viewpoint difference when trackers behave differently in the two views.

\begin{figure}[t]
    \centering
    \includegraphics[width=.8\columnwidth]{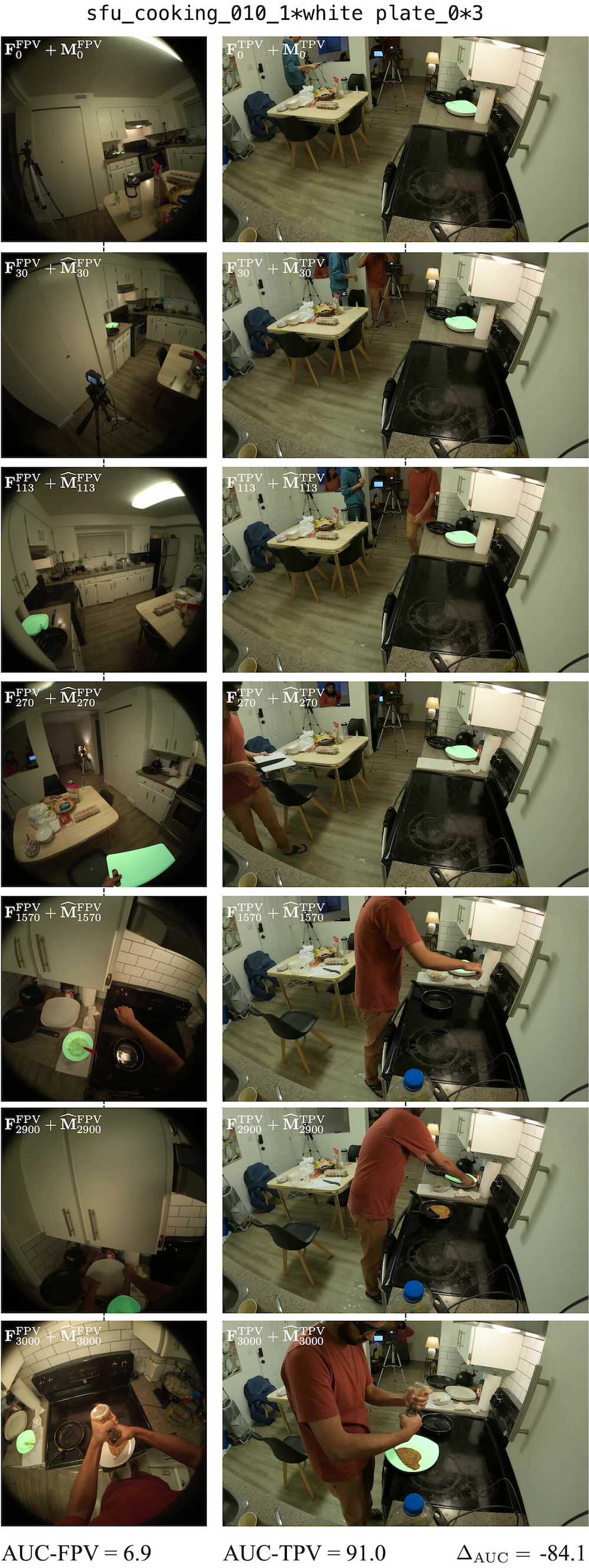}
    \caption{\textbf{Qualitative example \# 1.} Here, we illustrate the behavior of DAM4SAM \cite{dam4sam} on a sequence from \benchmarkname's evaluation set $\testset$. The frames $\frame_t^{\pov}$ are overlaid with the predicted segmentation masks $\widehat{\mask}_t^{\pov}$ (shown in light green). Below the frames, we report the sequence-wise AUC-FPV, AUC-TPV, and $\Delta_{\text{AUC}}$. In this example, the tracker loses the target early in the FPV sequence, whereas it remains stable in TPV despite object displacement. This discrepancy is reflected in the mean signed difference, which is highly negative.}
    \label{fig:qual1}
\end{figure}

\begin{figure}[t]
    \centering
    \includegraphics[width=.8\columnwidth]{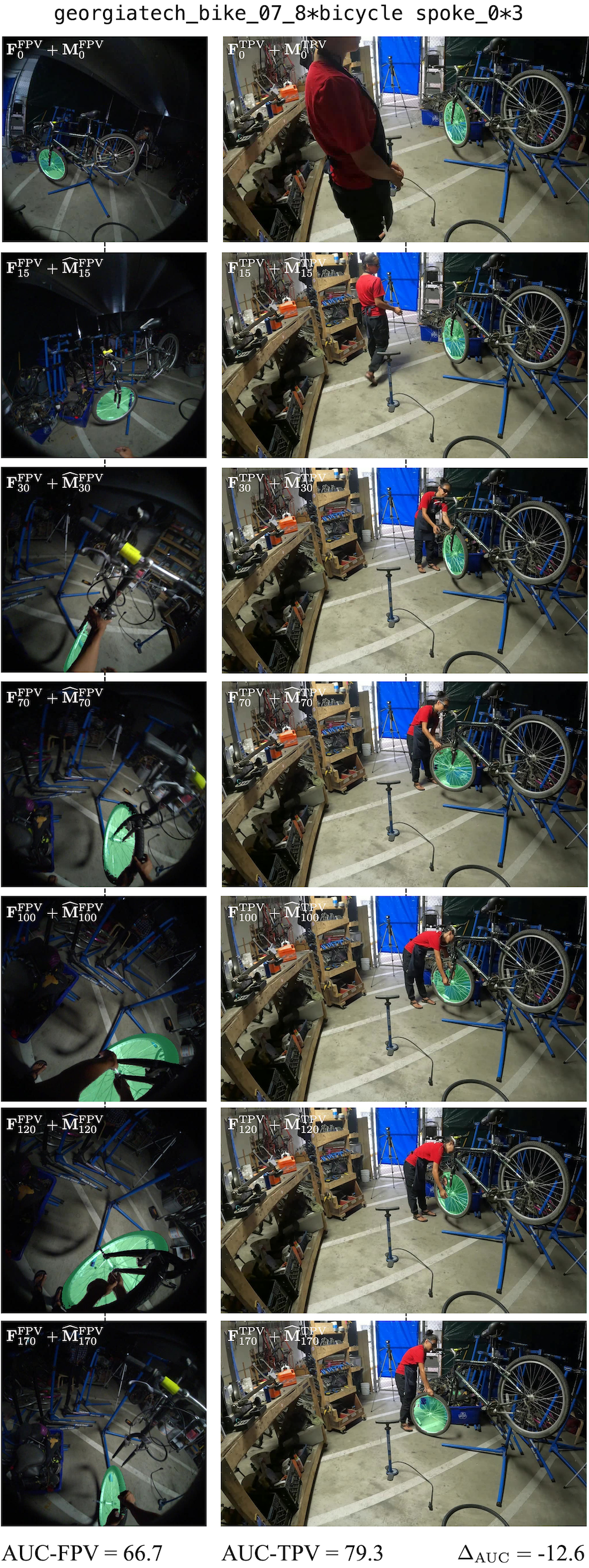}
    \caption{\textbf{Qualitative example \# 2.} Here, we illustrate the behavior of DAM4SAM \cite{dam4sam} on a sequence from \benchmarkname's evaluation set $\testset$. The frames $\frame_t^{\pov}$ are overlaid with the predicted segmentation masks $\widehat{\mask}_t^{\pov}$ (shown in light green). Below the frames, we report the sequence-wise AUC-FPV, AUC-TPV, and $\Delta_{\text{AUC}}$. In this example, the tracker remains relatively stable in both FPV and TPV. However, the high variability of the target appearance in FPV makes segmentation prediction more challenging. This difference is reflected in the mean signed difference, indicating better performance in TPV.}
    \label{fig:qual3}
\end{figure}

\begin{figure}[t]
    \centering
    \includegraphics[width=.8\columnwidth]{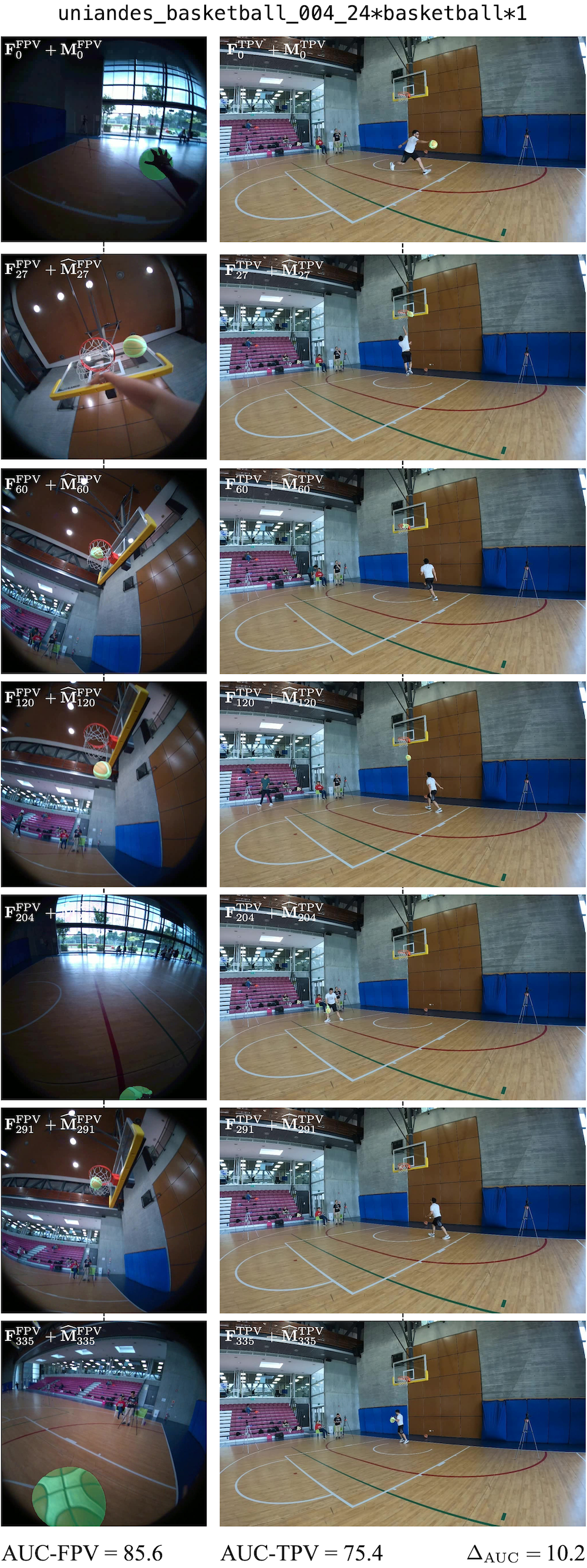}
    \caption{\textbf{Qualitative example \# 3.} Here, we illustrate the behavior of DAM4SAM \cite{dam4sam} on a sequence from \benchmarkname's evaluation set $\testset$. The frames $\frame_t^{\pov}$ are overlaid with the predicted segmentation masks $\widehat{\mask}_t^{\pov}$ (shown in light green). Below the frames, we report the sequence-wise AUC-FPV, AUC-TPV, and $\Delta_{\text{AUC}}$. In this example, the tracker remains relatively stable in both FPV and TPV. However, the lower resolution of the target in TPV makes segmentation prediction more challenging. This difference is reflected in the mean signed difference, indicating better performance in FPV.}
    \label{fig:qual2}
\end{figure}

\begin{figure}[t]
    \centering
    \includegraphics[width=.8\columnwidth]{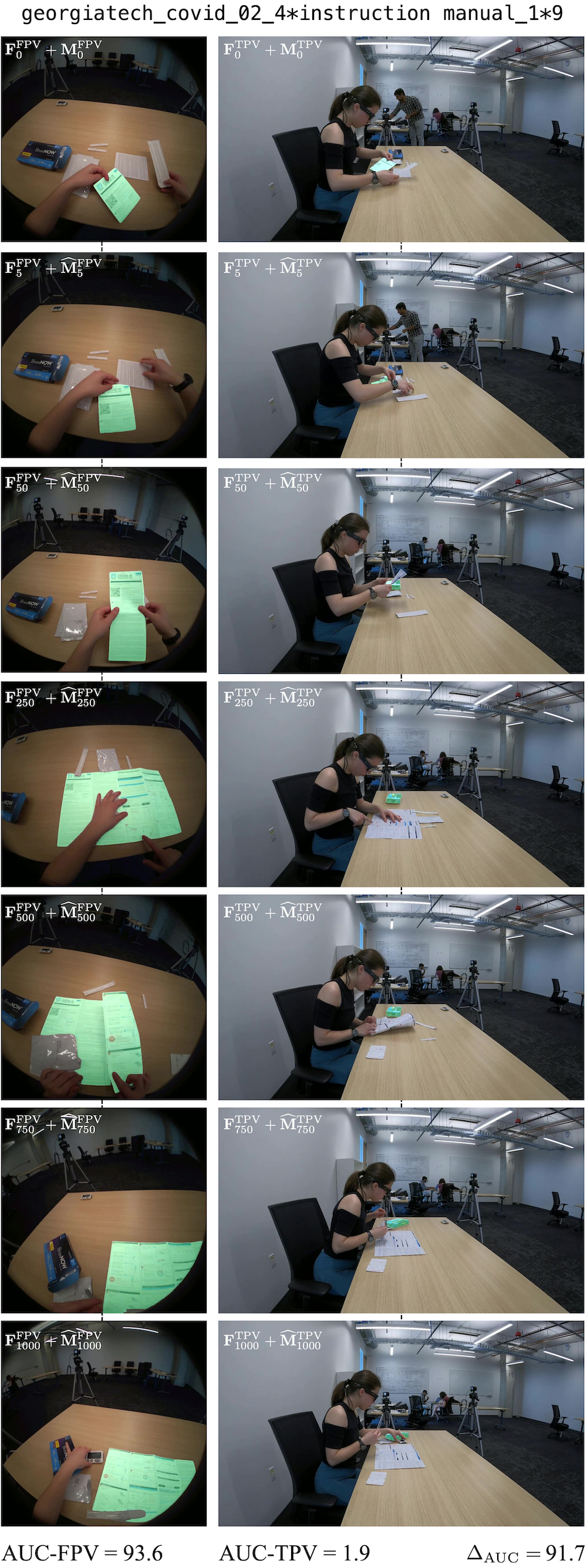}
    \caption{\textbf{Qualitative example \# 4.} Here, we illustrate the behavior of DAM4SAM \cite{dam4sam} on a sequence from \benchmarkname's evaluation set $\testset$. The frames $\frame_t^{\pov}$ are overlaid with the predicted segmentation masks $\widehat{\mask}_t^{\pov}$ (shown in light green). Below the frames, we report the sequence-wise AUC-FPV, AUC-TPV, and $\Delta_{\text{AUC}}$. In this example, the tracker loses the target early in the TPV sequence, whereas it remains stable in FPV despite object transformation. This discrepancy is reflected in the mean signed difference, which is highly positive.}
    \label{fig:qual4}
\end{figure}

\paragraph{Qualitative Results.} Fig. \ref{fig:qual1}, \ref{fig:qual3}, \ref{fig:qual2}, and \ref{fig:qual4} present qualitative examples of the most accurate tracker, DAM4SAM \cite{dam4sam}, on selected FPV and TPV sequences from the \benchmarkname\ test set. Each figure displays the predicted target segmentations alongside the sequence-wise AUC-FPV, AUC-TPV, and their signed difference.

\section{Limitations}

This study did not control the placement of TPV cameras. Although we selected a TPV view that provides the best visualization of the scene and activity \cite{EgoExo4D}, future work could compare FPV with multiple TPV positions to assess their impact on tracking performance and explore how different TPV configurations relate to FPV.

This study did not assess the impact of FPV and TPV tracking on downstream tasks. Our focus was to evaluate performance differences between FPV and TPV in the VOTS task \cite{EgoExo4D}. While prior work \cite{dunnhofer2023visual,manigrasso2024online} has shown a connection between tracking behavior and downstream task accuracy, we specifically examined object tracking performance. Future research could explore how the conclusions drawn from \benchmarkname\ relate to the performance of higher-level vision modules that rely on FPV or TPV object tracking.

\end{document}